\documentclass[10pt,twocolumn,letterpaper]{article}

\usepackage{cvpr}
\usepackage{times}
\usepackage{epsfig}
\usepackage{graphicx}
\usepackage{amsmath}
\usepackage{amssymb}
\usepackage{url}
\usepackage{subcaption}
\usepackage{paralist, tabularx}
\usepackage{booktabs}
\usepackage{multirow}
\usepackage{array}
\usepackage{makecell}
\usepackage{textcomp}
\usepackage{pifont}
\usepackage{wrapfig,lipsum}
\usepackage{titling}

\newcommand{\cmark}{\ding{51}}%
\newcommand{\xmark}{\ding{55}}%
\DeclareMathOperator*{\argmax}{arg\,max} 

\usepackage[pagebackref=true,breaklinks=true,letterpaper=true,colorlinks,bookmarks=false]{hyperref}

\cvprfinalcopy 

\ifcvprfinal\fi
\begin{document}

\title{Elucidating image-to-set prediction: \\ An analysis of models, losses and datasets}

\definecolor{brandeisblue}{rgb}{0.0, 0.44, 1.0}
\author{Luis Pineda$^{1}$\thanks{Equal contribution.}, Amaia Salvador$^{2}$\footnotemark[1]\,\,\thanks{Work partially done during internship at Facebook AI Research.}, Michal Drozdzal$^{1}$, Adriana Romero$^{1}$ \\
$^1$Facebook AI Research \:\:\:
$^2$Universitat Politecnica de Catalunya \\
\texttt{amaia.salvador@upc.edu, \{lep, mdrozdzal, adrianars\}@fb.com }
}
\date{}

\maketitle

\begin{abstract}

In this paper, we identify an important reproducibility challenge in the image-to-set prediction literature that impedes proper comparisons among published methods, namely, researchers use different evaluation protocols to assess their contributions. To alleviate this issue, we introduce an image-to-set prediction benchmark suite built on top of five public datasets of increasing task complexity that are suitable for multi-label classification (VOC, COCO, NUS-WIDE, ADE20k and Recipe1M). Using the benchmark, we provide an in-depth analysis where we study the key components of current models, namely the choice of the image representation backbone as well as the set predictor design. Our results show that (1) exploiting better image representation backbones leads to higher performance boosts than enhancing set predictors, and (2) modeling both the label co-occurrences and ordering has a slight positive impact in terms of performance, whereas explicit cardinality prediction only helps when training on complex datasets, such as Recipe1M. To facilitate future image-to-set prediction research, we make the code, best models and dataset splits publicly available at:  {\small\url{https://github.com/facebookresearch/image-to-set}}.
\end{abstract}

\section{Introduction}

Major advances in image understanding tasks \cite{vgg,resnet,faster-rcnn,yolo,long2015fully,jegou2017one,maskrcnn} have been enabled by the introduction of large scale datasets such as ImageNet~\cite{imagenet}, MS-COCO~\cite{lin2014microsoft} or Cityscapes~\cite{cityscapes}. All of these datasets come with benchmark suites that target well defined problems, provide dataset splits, and automated evaluation servers to rank methods according to their test set results, thus ensuring a fair comparison among methods. However, building large scale datasets with benchmark suites requires significant effort, which may not scale well with the increasing number of image understanding tasks and modalities. As a result, researchers often resort to reusing publicly available datasets, without defining rigorous benchmark suites for the task at hand. This lack of rigour leads to contributions with apparent new state-of-the-art results, which are evaluated under different dataset splits, evaluation metrics, experimental budgets~\cite{Melis2017}, and other uncontrolled sources of variations~\cite{Melis2017,Bouthillier19a}. As such, these seemingly successful empirical outcomes result in unquantifiable progress, raising reproducibility issues and driving potentially inaccurate claims. In the quest to draw sound and robust conclusions, we join other recent papers in highlighting the under-acknowledged existence of conclusion replication failure~\cite{Oliver2018,Locatello2018,Lucic2018,Kurach2018,Henderson2017}.

\begin{table*}[t!]
\centering
\resizebox{0.8\textwidth}{!}{
\begin{tabular}{@{}lccccccc@{}}
\toprule
\textbf{Set predictor}                     &\textbf{Backbone} & \textbf{Finetuned} &  \textbf{Train} & \textbf{Validation} & \textbf{Test} & \textbf{Baseline O-F1} & \textbf{O-F1} \\ \midrule
Li et al. 2017 \cite{li2017improving}        &  VGG  & \cmark & 77977 & 4104 & 40137 & n/a & 62.9* \\
Wang et al. 2017 \cite{wang2017multi}        &  VGG  & \cmark &  82081 & n/a & 40137 & n/a & 72.0* \\
Zhang et al. 2018 \cite{zhang2018multilabel} &  VGG  & \cmark &  82081 & n/a & 40137 & n/a & 66.5* \\
Chen et al. 2018 \cite{chen2018recurrent}    &  VGG  & n/a    &  82081 & n/a & 40137 & n/a  & 71.1\\
Liu et al. 2018  \cite{liu2018multi}         &  VGG  & \cmark &  82081 & n/a & 40137 & n/a  & 74.0\\
Wang et al. 2016 \cite{WangYMHHX16cnnrnn}  &    VGG  & \xmark &  82783 & n/a & 40504 & 63.3*  & 67.8*                \\
Rezatofighi et al. 2017 \cite{deepsetnet}  &    VGG     & \cmark & 74505 & 8278 & 40504 & 62.9* & 69.4                 \\
Liu et al. 2017 \cite{liu2017semantic}     &    VGG     & \cmark & 82783 & n/a & 40504 & 63.3 & 75.16                \\
Li et al. 2018 \cite{li2018attentive}     &     VGG     & \cmark & 82783 & n/a & 40504 & 59.3 & 65.2               \\
Rezatofighi et al. 2018 \cite{rezatofighi2017joint} & VGG& \cmark& 74505 & 8278 & 40504 & 69.2* & 70.7                \\
Luo et al. 2019 \cite{luo2019visual}         & ResNet-50  & \cmark & n/a & n/a & n/a & 63.2 & 65.2
\\
Ge et al. 2018 \cite{ge2018multi}            & ResNet-101 & \cmark & 82081 & n/a & 40504 & 76.3 & 78.4
\\
Zhu et al. 2017 \cite{zhu2017learning}     & ResNet-101 & \cmark & 82783 & n/a & 40504 & 74.4 & 75.8                 \\
Guo et al. 2019 \cite{guo2019visual}      & ResNet-101  & \cmark & 82783 & n/a & 40504 & 73.7 & 76.3 \\
Liu et al. 2019 \cite{liu2019decoupling}  & ResNet-101  & n/a    & 82787 & n/a & 40504 & 77.1 & 79.5 \\
Chen et al. 2019 \cite{chen2019multi}     & ResNet-101  & n/a    & 82081 & n/a & 40504 & 76.8 & 80.3 \\
Chen et al. 2018 \cite{chen2018order}      & ResNet-152 & \xmark & 82783 & n/a & 40504 & 61.0 & 67.7                \\
\bottomrule
\end{tabular}}
\caption{\textbf{Overview of image-to-set prediction methods applied to MS-COCO.} Backbone refers to the pre-trained image representation model, whether finetuned or not (n/a indicates the lack of finetuning information). Train, validation and test indicate the number of images used in each split. When validation is n/a, the same split has been used for both validation and test. Baseline O-F1 corresponds to the (reported) results of training each method's backbone with binary cross-entropy. O-F1 corresponds to each method's best reported result. Note that * refers to results which limit the cardinality of predictions to $3$ or $4$ elements. }
\label{tab:sota_coco}
\end{table*}

In particular, we build a case study for an important computer vision problem, the image-to-set prediction task (also referred to as multi-label classification), as everyday life pictures are typically complex scenes which can be described with multiple concepts/objects. The most widely used large scale dataset to assess the progress of image-to-set prediction is the MS COCO~\cite{lin2014microsoft} object detection dataset. Table \ref{tab:sota_coco} summarizes recent contributions in the image-to-set prediction literature, emphasizing their factors of variation, and displaying their reported results. As shown in the table, the discordance across published methods is surprisingly high, hindering the robustness of method comparison. First, we notice factors of variation that arise from the lack of a well established benchmark suite: (1) different methods use different train, validation and test splits, (2) not all approaches finetune the image representation backbone and (3) some approaches artificially limit the cardinality of the predicted sets. Notably, when validation set information is unavailable (n/a), it is unclear how model selection is performed. However, in cases where code is made publicly available, one can notice that the test set is used either for hyperparameter selection or for model early stopping. Second, we highlight additional factors of variations due to advancements in image classification, namely the choice of image representation backbone. Finally, we draw the reader's attention to the results reported for a simple baseline model trained with binary cross-entropy which exhibits surprisingly high variance (e.g. for ResNet-101, the best reported baseline score is $77.1\%$, while the worst reported value for the same model is $73.7\%$). All these discrepancies raise the question of fair comparison across models, hampering the conclusions about the role of individual model components in image-to-set prediction advancement.~\footnote{Similar observations can be made for other datasets, e.g. in case of NUS-WIDE dataset, the number of images used by different works varies since the images are downloaded at different times and some download links are inactive \cite{liu2017semantic}. Moreover, even when ensuring the same number of images, the dataset splits are defined randomly \cite{li2017improving,liu2017semantic,zhu2017learning}.}

Therefore, in this paper, we argue that enabling the community with a proper benchmark suite is of crucial importance to take firm steps towards advancing image-to-set prediction methods. The proposed benchmark suite is comprised of a unified code-base, including dataset splits for 5 datasets of increasing complexity (Pascal VOC 2007 \cite{everingham2010pascal}, MS COCO 2014 \cite{lin2014microsoft}, ADE20k \cite{zhou2017scene}, NUS-WIDE \cite{chua2009nus} and Recipe1M \cite{salvador2017learning}) and a common evaluation protocol designed to assess the impact of architecture innovations, ensure reproducibility of results and, perhaps more importantly, strengthen the robustness of conclusions. Together with the benchmark suite, we provide an extensive study to weigh the influence of prominent innovations and baselines in the image-to-set literature. Moreover, to ensure that differences in model performance can be attributed to modeling choices, rather than unbalanced hyperparameter search, we use a fixed budget of tested configurations (allowing all models to have equal opportunity to reach their best results) by means of the \textsc{Hyperband} algorithm~\cite{li2018hyperband}.  
Our analysis aims to investigate the importance of key image-to-set prediction model components: the image representation backbone and the set predictor. On the one hand, we are interested in understanding whether architectural improvements from the single-class image classification literature translate into the multi-label classification scenario. On the other hand, we aim to analyze the importance of (1) explicitly modeling label co-occurrences, (2) leveraging label ordering, and (3) including cardinality prediction in the set predictor. 
Our main observations can be summarized as:
\begin{compactitem}
   \item[--] Image-to-set prediction benefits from architectural improvements in the single label classification literature.
   \item[--] Explicitly leveraging label co-occurrences and label ordering tends to have a slight positive impact in terms of performance, whereas incorporating cardinality prediction only helps when training on complex datasets, such as Recipe1M.
   \item[--] Exploiting  better  image  representation backbones tends to lead to higher performance boosts than enhancing set predictors. In particular, simple baselines such as training image representation backbones with binary cross-entropy have the potential to outperform other methods when combined with recent architectural advancements for single label classification and given enough hyper-parameter search budget.
\end{compactitem}

\section{Overview of multi-label classification}

Multi-label classification has been a long lasting problem in computer vision \cite{Zhang2007,NIPS2009_3824,Zhao2015,WuJLGL18} tackled by a wide variety of methods, such as, decomposing the problem into independent single-label classification problems \cite{Nam2014,Zhang2007}, exploiting label co-occurrences \cite{AntonucciCMG13,ShuLXT15,Liu2015LargeMM}, introducing priors such as label noise and sparsity \cite{NIPS2009_3824,NIPS2012_4591,SongMVK15,WuJLGL18,Zhao2015,Bi2013}, and more recently, by leveraging deep neural networks \cite{yang2016exploit,chen2016deep,WangYMHHX16cnnrnn}. Approaches in the deep learning realm have also attempted to decompose the multi-label classification problem into single-label classification problems, by independently classifying features extracted from object proposals \cite{yang2016exploit, wei2016hcp,liu2018multi, ge2018multi} or by considering global image features and finetuning pre-trained models with a binary logistic loss \cite{chen2016deep, chatfield2014return, zhu2017learning, guo2019visual, liu2019decoupling}. In order to explicitly exploit label co-occurrences, researchers have resorted to modeling the joint probability distribution of labels \cite{Tsoumakas10powerset} or decomposing the joint distribution into conditionals \cite{Dembczynski2010pcc,WangYMHHX16cnnrnn,Nam17setrnn,li2018attentive, liu2017semantic, lyu2019attend, nauata2018structured}. This has been done, for example,  by using recurrent neural networks, at the expense of introducing intrinsic label ordering during training, which has been resolved by applying a category-wise max-pooling across the time dimension \cite{wang2017multi, chen2018recurrent, zhang2018multilabel,inversecooking} or by optimizing for the most likely ground truth label at each time step \cite{chen2018order}. Alternative solutions to capture label co-occurrences include learning joint input-label embeddings with ranking-based losses \cite{Weston2011WSU, Lin2014MCV, YehWKW17, li2017improving, GongJLTI13}, graph neural networks \cite{chen2019multi}, and using loss functions that directly account for those \cite{GongJLTI13,WeiXHNDZY14,Mahajan18hashtags,inversecooking}. Finally, multi-label classification has only recently been posed as a set prediction problem, where both set elements (labels) and cardinality are predicted, e.g. \cite{deepsetnet,rezatofighi2017joint} model cardinality as a categorical distribution, \cite{li2017improving} learns class-specific probability thresholds, and \cite{NIPS2018_7820} frames set prediction as a parameterized policy search problem.

\section{Benchmark methodology}
\label{sec:archs}

\begin{figure*}[t!]
    \begin{subfigure}{0.45\textwidth}
    \vspace{-1.8cm}
    \resizebox{\columnwidth}{!}{
\begin{tabular}{@{}ccccc@{}}
\toprule
Model & loss & co-occurrences & cardinality & ordering\\ \midrule
$\mathrm{FF}$ & BCE & \xmark & \xmark & \xmark\\
$\mathrm{FF}$ & sIoU & $\mathcal{L}$ & \xmark & \xmark\\
$\mathrm{FF}$ & TD & $\mathcal{L}$ & \xmark & \xmark\\ \midrule
$\mathrm{FF}$ & BCE & \xmark & DC dist. & \xmark\\
$\mathrm{FF}$ & BCE & \xmark & C dist.  & \xmark\\
$\mathrm{FF}$ & sIoU & $\mathcal{L}$ & C dist.  & \xmark\\
$\mathrm{FF}$ & TD & $\mathcal{L}$ & C dist.  & \xmark\\ \midrule
$\mathrm{LSTM}$& CE & $\theta$ & \emph{eos} token & \cmark\\
$\mathrm{LSTM}_{set}$ & BCE & $\theta$ & \emph{eos} token & \xmark\\
$\mathrm{TF}$ & CE & $\theta$ & \emph{eos} token & \cmark\\
$\mathrm{TF}_{set}$ & BCE & $\theta$ & \emph{eos} token & \xmark\\
\bottomrule
\end{tabular}}
\vspace{.2cm}
\caption{}
\label{tab:models_summary}
\vspace{2cm}
\end{subfigure}\hfill
    \begin{subfigure}[t]{0.16\textwidth}
        \includegraphics[width=\textwidth]{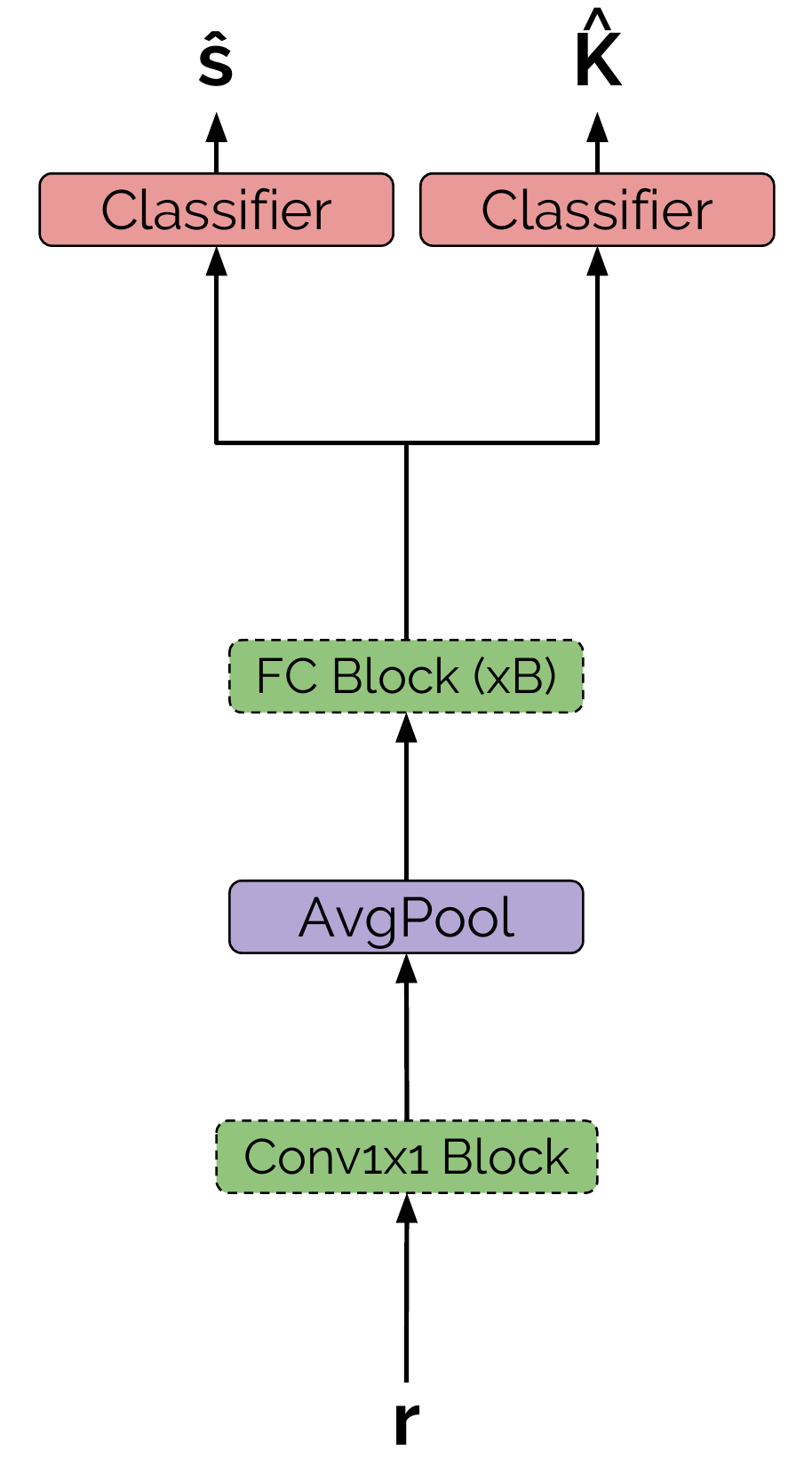}
        \vspace{.02cm}
        \caption{}
        \label{fig:fc}
    \end{subfigure}\hfill
        \begin{subfigure}[t]{0.16\textwidth}
        \includegraphics[width=\textwidth]{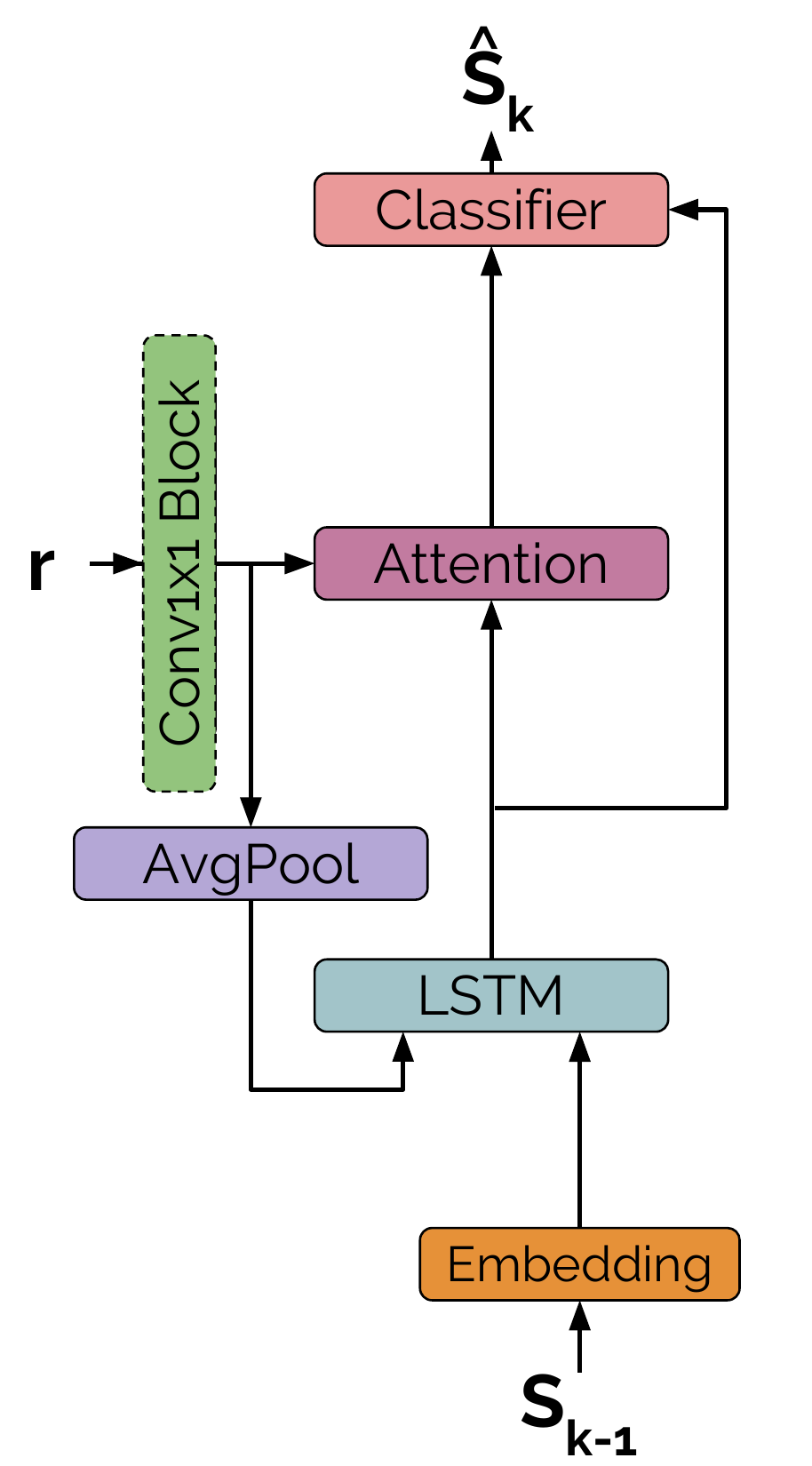}
        \vspace{.02cm}
        \caption{}
        \label{fig:lstm}
    \end{subfigure}\hfill
    \begin{subfigure}[t]{0.16\textwidth}
        \includegraphics[width=\textwidth]{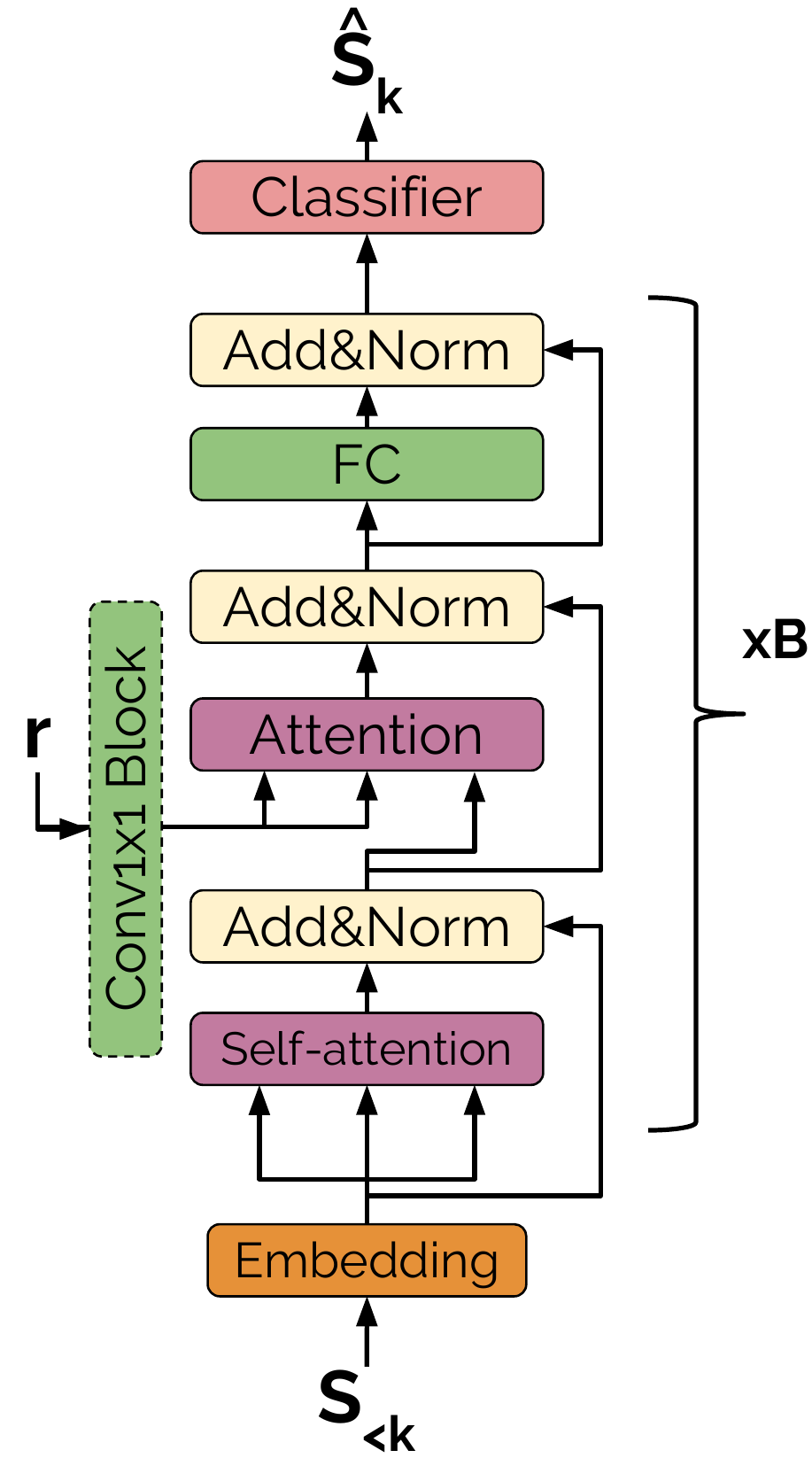}
        \vspace{.02cm}
        \caption{}
        \label{fig:tf}
    \end{subfigure}
    \vspace{-2cm}
\caption{\textbf{(a) Models summary:} Loss-based modeling of label co-ocurrences is denoted with $\mathcal{L}$, while explicitly modeling dependencies through architecture design is represented by $\theta$. Notation: FF (feed-forward), LSTM (long short-term memory), TF (transformer), BCE (binary cross-entropy), sIoU (soft intersection-over-union), TD (target distribution), CE (categorical cross-entropy), DC dist. (Dirichlet-Categorical) and C dist. (Categorical distribution). \textbf{(b--d) Set prediction architectures: }(b) Feed-forward (FF), (c) LSTM \cite{lu2017knowing} and (d) Transformer \cite{transformer}.}
\label{fig:model_architectures}
\end{figure*}

In this section, we introduce our benchmark methodology. We start by reviewing the image-to-set prediction models, and follow by thoroughly describing the adopted hyper-parameter search strategy as well as the evaluation metric.

\subsection{Image-to-set prediction models}
\label{sec:i2s}
In image-to-set prediction, we are given a dataset of image and set of labels pairs, with the goal of learning to produce the correct set of labels for a given image. The set of labels is an unordered collection of unique elements, which may have variable size. Let $\mathcal{D} = \{d_i\}_{i=1}^{N}$ be a dictionary of labels of size $N$, from which we can obtain the set of labels $S$ for an image $\mathbf{x}$ by selecting $K \geq 0$ elements from $\mathcal{D}$. If $K = 0$, no elements are selected and $S = \{\}$; otherwise $S = \{s_i\}_{i=1}^{K}$. Thus, our training data consists of $M$ image and label pairs $\{(\mathbf{x}^{(i)}, S^{(i)})\}_{i=1}^{M}$. 

Image-to-set prediction models are composed of an \emph{image representation} backbone, followed by a \emph{set prediction} module, which are stacked together and trained end-to-end. Image representation backbones transform an input image $\mathbf{x} \in \mathbb{R}^{W \times H \times 3}$ into a representation $\mathbf{r} = f_{\phi}(\mathbf{x}) \in \mathbb{R}^{w \times h \times 2\,048}$, where $W \times H$ and $w \times h$ are the spatial resolutions of the image and its extracted features. Set prediction module takes as an input the image representation and outputs set elements. As \emph{image representation} backbone, we choose the output of the last convolutional layer of a top performing convolutional network pre-trained on ImageNet \cite{imagenet}. In particular, we choose among popular image classification CNN architectures (ResNet-50 \cite{resnet}, ResNet-101 \cite{resnet} and ResNeXt-101-32x8d \cite{xie2016groups}) to assess whether multi-label classification can also benefit from improvements in the single image classification literature. As \emph{set predictor}, we consider feed-forward and auto-regressive architectures. A comprehensive overview of set predictor modules is out of the scope of this paper. We limit the scope of our study to assess the importance of design choices that (1) exploit \emph{label co-occurrences} (either explicitly through model design or through the loss function), (2) leverage \emph{label ordering} (e.g. auto-regressive models) and, (3) predict \emph{set cardinality} as part of their pipeline (either through a categorical output, or an end-of-sequence -- \emph{eos} -- token). Figure \ref{tab:models_summary} summarizes the image-to-set prediction models considered in this study.

\subsubsection{Feed-forward Set Predictors}
\label{ssec:ffmodels}

\textbf{Notation.} We represent $S$ as a binary vector $\mathbf{s}$ of dimension $N$, where $\mathbf{s}_i=1$ if $\mathbf{s}_i \in S$ and $0$ otherwise. The goal is to estimate the label probabilities $\mathbf{\hat{s}}$ from an image $\mathbf{x}$. 

\noindent \textbf{Architectures.} Feed-forward models take image features $\mathbf{r}$ as input and output $\mathbf{\hat{s}} = g_{\theta}(\mathbf{r})$. These models are composed of (1) an optional $1\times1$ convolutional block to change the feature dimensionality of the input, (2) a global average pooling operation to collapse the spatial dimensions, and (3) one or more fully connected layers. Intermediate fully connected layers are followed by dropout, batch normalization and a ReLU non-linearity. The last fully connected layer serves as classifier, and thus, is followed by a either a sigmoid or softmax non-linearity. The architecture used for all feed-forward models is depicted in Figure \ref{fig:fc}. 

\noindent \textbf{Loss functions.} 
The model's parameters are trained by maximizing:
\begin{equation}
   \argmax_{\phi, \theta} \sum_{i=0}^M \log p(\mathbf{\hat{s}}^{(i)}=\mathbf{s}^{(i)}|\mathbf{x}^{(i)};\phi, \theta).
\label{eq:ing_set}
\end{equation}
where $\phi$ and $\theta$ are the image representation and set predictor parameters, respectively. Most state-of-the-art feed-forward methods assume independence among labels, factorizing $\log p(\mathbf{\hat{s}}^{(i)}=\mathbf{s}^{(i)}|\mathbf{x}^{(i)})$ as $\sum_{j=0}^N \log p(\mathbf{\hat{s}}_j^{(i)}=\mathbf{s}_j^{(i)}|\mathbf{x}^{(i)})$ and using binary cross-entropy (BCE) as training loss. However, the elements in the set are not necessarily independent. In order to account for label co-occurrences, we borrow from the semantic segmentation literature and train the feed-forward set predictor with a soft structured prediction loss, such as the \emph{soft intersection-over-union} (sIoU) \cite{Drozdzal2016}.
Alternatively, we use the \emph{target distribution} (TD) \cite{GongJLTI13,Mahajan18hashtags} to model the joint distribution of set elements and train a model by minimizing the cross-entropy loss between $p(\mathbf{s}^{(i)}|\mathbf{x}^{(i)}) = \mathbf{s}^{(i)}/{\sum_j \mathbf{s}_j^{(i)}}$ and the model's output distribution $p(\mathbf{\hat{s}}^{(i)}|\mathbf{x}^{(i)})$. We refer to the feed-forward models trained with the aforementioned losses as $\mathrm{FF_{BCE}}$, $\mathrm{FF_{sIoU}}$, and $\mathrm{FF_{TD}}$, respectively. Note that, by construction, none of these models exploit label ordering.

\noindent \textbf{Set cardinality.} 
Given the probabilities $\mathbf{\hat{s}}$ estimated by a feed forward model, a set of labels $\hat{S}$ must be recovered. For $\mathrm{FF_{BCE}}$ and $\mathrm{FF_{sIoU}}$, one simple solution is to apply a threshold $t$ to $\mathbf{\hat{s}}$, keeping all labels for which $\mathbf{\hat{s}}_i \geq t$. Typically, this threshold is set to $0.5$. Nonetheless, in the case of the $\mathrm{FF_{TD}}$, we adopt the strategy of \cite{inversecooking} and recover the label set by greedily sampling elements from a \emph{cumulative distribution of sorted output probabilities} $p(\mathbf{\hat{s}}^{(i)}|\mathbf{x}^{(i)})$. We stop the sampling once the sum of probabilities of selected elements is $> 0.5$. Alternatively, the set cardinality $K$ may be explicitly predicted by the feed-forward model through a second output $\{\mathbf{\hat{s}},\mathbf{\hat{K}}\}=g_{\theta}(\mathbf{r})$, where $\mathbf{\hat{K}}$ estimates the probabilities over possible set cardinalities. At inference time, the top-$\hat{K}$ labels are included in the predicted set. We refer to these models as $\mathrm{FF_{BCE, C}}$. For completeness, we also consider a variant of $\mathrm{FF_{BCE}}$ where the set cardinality is modeled with a Dirichlet-Categorial distribution ($\mathrm{FF_{BCE, DC}}$), following \cite{rezatofighi2017joint}.

\noindent \textbf{Empty set prediction.}
Unlabeled images can be naturally handled by models, whose output estimates a probability per label (e.g. $\mathrm{FF_{BCE}}$ and $\mathrm{FF_{sIoU}}$). At inference time, the set cardinality is determined by applying a threshold $t$ to each output probability. The set cardinality can also be explicitly predicted by a feed-forward model through a second output, where the output of cardinality $0$ corresponds to empty set. From the feed-forward models considered, only  $\mathrm{FF_{TD}}$ cannot handle empty sets, since a vector with all zeros is not a valid (categorical) probability distribution.

\subsubsection{Auto-regressive Set Predictors}
\label{ssec:autoreg}
\textbf{Notation.} We represent $S$ as a $K \times N$ binary matrix $\mathbf{S}$.
We set $\mathbf{S}_{i,j}=1$ if label $d_j$ is selected at $i$-th position and $0$ otherwise. Each row in $\mathbf{S}$ contains the one-hot-code representation of one label.

\noindent \textbf{Architectures.} We explore two auto-regressive architectures: a Long Short-Term Memory ($\mathrm{LSTM}$)~\cite{lstm} with spatial attention-based model \cite{lu2017knowing} and a transformer-based one ($\mathrm{TF}$)~\cite{transformer}. Both architectures take image features $\mathbf{r}$ as input and output $\mathbf{\hat{S}} = g_{\theta}(\mathbf{r})$. These models are composed of either a single LSTM layer -- following \cite{lu2017knowing} --, or several transformer layers -- following \cite{transformer} and \cite{inversecooking}. 
The output layer of the model is used as classifier and has a softmax non-linearity. These models sequentially predict set labels. Their architectures are depicted in Figures \ref{fig:lstm} and \ref{fig:tf}.

\noindent \textbf{Loss functions.} The models' parameters are trained to predict $\mathbf{\hat{S}}$ from an image $\mathbf{x}$ by maximizing:
\begin{equation}
   \argmax_{\phi, \theta} \sum_{i=0}^M \log p(\mathbf{\hat{S}}^{(i)}=\mathbf{S}^{(i)}|\mathbf{x}^{(i)};\phi, \theta).
\label{eq:ing_list}
\end{equation}
To ensure that labels in $\mathbf{\hat{S}}^{(i)}$ are selected without repetition, we force the pre-activation of $p(\mathbf{\hat{S}}_{k}^{(i)}|\mathbf{x}^{(i)}, \mathbf{S}_{<k}^{(i)})$ to be $-\infty $ for all previously selected labels. One characteristic of the formulation in Equation \ref{eq:ing_list} is that it inherently exploits label ordering, which might not necessarily be relevant for the set prediction task. In order to ignore the order in which labels are predicted, we employ the solution of \cite{wang2017multi}, \cite{chen2018recurrent} and \cite{inversecooking}, and aggregate the outputs across different time-steps by means of a max pooling operation. In this case, instead of minimizing the cross-entropy error at each time step, we minimize the BCE between the pooled predicted labels and the ground truth. We refer to the $\mathrm{LSTM}$ and $\mathrm{TF}$ models trained with pooled time-steps as $\mathrm{LSTM_{set}}$ and $\mathrm{TF_{set}}$, respectively. It is worth noting that, in all cases, at inference time, we directly sample from the auto-regressive predictor's output. As an alternative to prevent auto-regressive models from exploiting label ordering, we also consider a variant where we randomly shuffle the label ordering of each sample ($\mathrm{LSTM_{shuffle}}$ and $\mathrm{TF_{shuffle}}$).

\noindent \textbf{Set cardinality.} Most auto-regressive set predictors in the literature are not concerned with cardinality prediction, and predict a fixed number of labels by default \cite{chen2018recurrent, WangYMHHX16cnnrnn}. However, we argue that those models inherently have the mechanism to learn when to stop. Therefore, as commonly done in tasks such as image captioning and machine translation, we incorporate an \emph{eos} token, which has to be predicted in the last sequence step. Thus, the \emph{eos} token's role is to estimate the cardinality of the set. 
In the case of $\mathrm{LSTM_{set}}$ and $\mathrm{TF_{set}}$, we learn the stopping criterion with an additional loss weighted by means of a hyperparameter $\lambda_{eos}$. The \emph{eos} loss is defined as the BCE between the predicted \emph{eos} probability at different time-steps and the ground truth. 

\begin{figure*}[t!]
\begin{minipage}{0.5\linewidth}
\resizebox{\columnwidth}{!}{
\begin{tabular}[b]{@{}lccccc@{}}
\toprule
      & VOC & COCO & NUS-WIDE & ADE20k & Recipe1M \\ \midrule
Train &     $4\,509$ &     $74\,503$    &  $145\,610$ & $18\,176$      &  $252\,547$        \\
Val   & $502$    &   $8\,280$      &  $16\,179$   & $2\,020$     &  $5\,000$        \\
Test  &   $4\,952$  &    $40\,504$     &    $107\,859$    & $2\,000$  &   $54\,506$       \\ \midrule
$N$  &   $20$  &    $80$     &    $81$    & $150$  &   $1\,486$  \\ \midrule
$K$  &   \makecell{$1.57$ \\ (0.77)}  &   \makecell{$2.91$ \\ (1.84)}     &    \makecell{$1.86$ \\ (1.71)} & \makecell{$8.17$ \\ (4.14)}     &   \makecell{$7.99$ \\ (3.21)}       \\\bottomrule
\end{tabular}}
\captionof{table}{\textbf{Dataset summary.} Splits, dictionary size ($N$), and cardinality ($K$), reported as $mean$ ($std$) for each dataset.}
\label{tab:dataset_splits}
\end{minipage}
    \quad
\begin{minipage}{0.5\linewidth}
    \centering
        \includegraphics[width=\columnwidth]{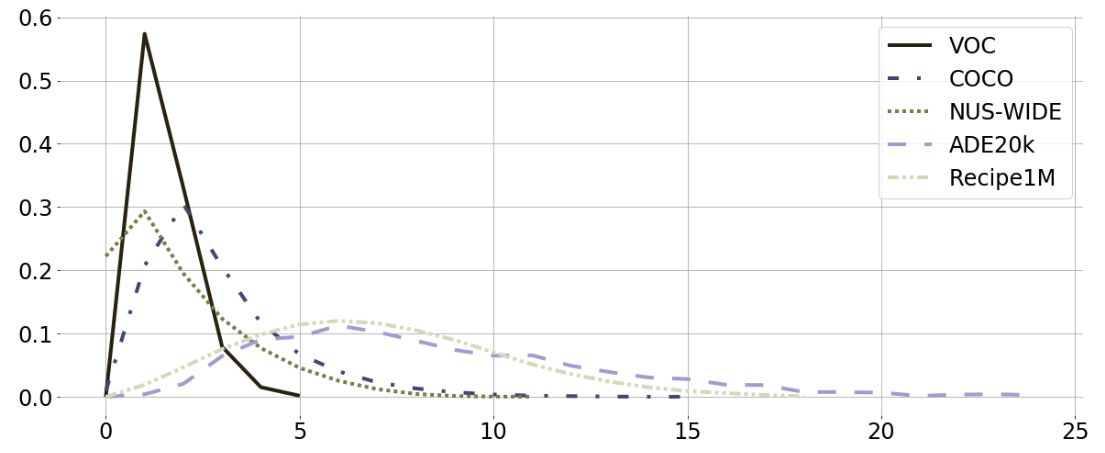}
        \captionof{figure}{\textbf{Dataset cardinality distribution.}}
        \label{fig:cardinality_distr}
\end{minipage}
\vspace{-0.35cm}
\end{figure*}

\noindent \textbf{Empty set prediction.}
We handle images with missing labels by setting the \emph{eos} token as the first element to be predicted in the sequence. 
\subsection{Hyper-parameter search}
\label{ssec:hp_search}

To warrant that differences in model performances can be attributed to modeling choices, we ensure that all models have an equal opportunity to shine by granting each one of them the same hyperparameter search budget through the \textsc{Hyperband}~\cite{li2018hyperband} algorithm. \textsc{Hyperband} is a bandit-based algorithm that speeds up random search via a more robust variant of the \textsc{SuccessiveHalving} early-stopping algorithm~\cite{jamieson2016non}. 
We have opted for \textsc{Hyperband} because it is extremely parallelizable, extensive evaluation has shown it achieves similar performance to more complex methods~\cite{li2018hyperband}, and it has theoretical guarantees that do not rely on strong assumptions about the function to be optimized (in our case best overall F1 validation set score). 

\subsection{Evaluation protocol}

We evaluate methods by means of F1 score that combines both precision and recall in a single score. For each tested model, we report three F1 values: per-class (C-F1), per-image (I-F1) and overall (O-F1).

\section{Experiments}

We ensure proper comparison among models by considering unified dataset splits and evaluate our models on 5 different datasets. Among those, we consider Pascal VOC 2007, MS COCO, and NUS-WIDE (following previous work but unifying dataset splits). Pascal VOC 2007 and MS COCO are object detection datasets and, as such, contain partially or fully visible objects exclusively. This is not the case of NUS-WIDE, which also contains concepts with higher degree of abstraction. However, all the aforementioned datasets have a rather limited dictionary size (below $100$) and a small number of annotations per image ($< 3$ on average). In the quest for pushing the boundaries of image-to-set prediction methods, we incorporate ADE20k \cite{zhou2017scene} and Recipe1M \cite{salvador2017learning} datasets to our analysis, since they exhibit significantly larger dictionary sizes ($150$ and $1496$ vs $<100$) and greater number of annotations per image ($8$ on average). Moreover, Recipe1M includes image annotations corresponding to invisible items. Please refer to Table \ref{tab:dataset_splits} and Figure \ref{fig:cardinality_distr} for additional dataset information. A detailed description of each dataset and \textsc{Hyperband} tuning parameters are provided in the supplementary material.

\subsection{Analysis of set predictors}
\begin{table*}[t!]
\centering
\resizebox{\textwidth}{!}{
\begin{tabular}{@{}llccc|ccc|ccc|ccc|ccc@{}}
\toprule
 & & \multicolumn{3}{c}{\textbf{VOC}}        & \multicolumn{3}{c}{\textbf{COCO}}          & \multicolumn{3}{c}{\textbf{NUS-WIDE}}      & \multicolumn{3}{c}{\textbf{ADE20k}}        & \multicolumn{3}{c}{\textbf{Recipe1M}}   \\
\textbf{Rank} &\textbf{Model}                       & \textbf{O-F1}      & \textbf{C-F1}     & \textbf{I-F1}     & \textbf{O-F1}      & \textbf{C-F1}     & \textbf{I-F1}     & \textbf{O-F1}      & \textbf{C-F1}     & \textbf{I-F1}      & \textbf{O-F1}      & \textbf{C-F1}     & \textbf{I-F1}    & \textbf{O-F1}      & \textbf{C-F1}     & \textbf{I-F1}    \\ \midrule

1 & $\mathrm{FF_{BCE}}$ & \makecell{86.40\\(0.19)} & \makecell{85.16\\(0.20)} & \makecell{88.23\\(0.15)} & \makecell{76.98\\(0.06)} & \makecell{73.49\\(0.14)} & \makecell{79.35\\(0.07)} & \textbf{\makecell{71.34\\(0.06)}} & \textbf{\makecell{54.94\\(0.48)}} & \makecell{69.19\\(0.12)} & \makecell{70.89\\(0.34)} & \makecell{47.87\\(1.28)} & \makecell{69.43\\(0.30)} & \makecell{46.83\\(0.07)} & \makecell{18.38\\(0.17)} & \makecell{43.63\\(0.07)}\\ \midrule
2 & $\mathrm{LSTM}$ & \makecell{86.36\\(0.18)} & \makecell{85.00\\(0.20)} & \makecell{88.20\\(0.19)} & \makecell{76.63\\(0.08)} & \makecell{72.98\\(0.07)} & \makecell{79.45\\(0.09)} & \makecell{70.85\\(0.07)} & \makecell{54.15\\(0.16)} & \textbf{\makecell{69.43\\(0.04)}} & \makecell{70.68\\(0.23)} & \textbf{\makecell{48.73\\(1.40)}} & \makecell{69.97\\(0.23)} & \makecell{47.33\\(0.05)} & \makecell{17.55\\(0.05)} & \makecell{46.12\\(0.06)}\\ \midrule
3 & $\mathrm{TF}$ & \makecell{85.89\\(0.16)} & \makecell{84.28\\(0.26)} & \makecell{87.87\\(0.16)} & \makecell{76.62\\(0.11)} & \makecell{73.32\\(0.12)} & \makecell{79.45\\(0.07)} & \makecell{70.30\\(0.06)} & \makecell{53.31\\(0.45)} & \makecell{69.08\\(0.07)} & \makecell{70.46\\(0.15)} & \makecell{48.03\\(0.51)} & \makecell{69.42\\(0.25)} & \makecell{47.77\\(0.07)} & \makecell{17.93\\(0.07)} & \makecell{46.58\\(0.08)}\\ \midrule
4 & $\mathrm{FF_{BCE,C}}$ & \makecell{84.59\\(0.14)} & \makecell{84.04\\(0.18)} & \makecell{86.74\\(0.18)} & \makecell{75.40\\(0.06)} & \makecell{72.23\\(0.13)} & \makecell{78.16\\(0.08)} & \makecell{69.61\\(0.05)} & \makecell{52.14\\(0.72)} & \makecell{67.63\\(0.29)} & \makecell{70.19\\(0.13)} & \makecell{44.15\\(0.47)} & \makecell{69.20\\(0.13)} & \textbf{\makecell{50.22\\(0.03)}} & \makecell{18.28\\(0.07)} & \textbf{\makecell{48.47\\(0.04)}}\\ \midrule
5 & $\mathrm{TF_{shuffle}}$ & \makecell{86.79\\(0.25)} & \makecell{85.62\\(0.18)} & \makecell{88.63\\(0.22)} & \makecell{77.04\\(0.05)} & \makecell{73.72\\(0.03)} & \makecell{79.99\\(0.04)} & \makecell{69.51\\(0.19)} & \makecell{52.74\\(0.25)} & \makecell{68.23\\(0.15)} & \makecell{70.26\\(0.25)} & \makecell{45.71\\(1.08)} & \makecell{69.47\\(0.24)} & \makecell{46.78\\(0.09)} & \makecell{18.94\\(0.12)} & \makecell{45.62\\(0.11)}\\ \midrule
6 & $\mathrm{FF_{TD,C}}$ & \makecell{84.77\\(0.10)} & \makecell{83.61\\(0.26)} & \makecell{87.03\\(0.08)} & \makecell{74.99\\(0.08)} & \makecell{71.90\\(0.14)} & \makecell{78.11\\(0.05)} & \makecell{69.09\\(0.10)} & \makecell{51.33\\(0.84)} & \makecell{67.52\\(0.36)} & \makecell{69.35\\(0.13)} & \makecell{48.12\\(0.27)} & \makecell{68.48\\(0.19)} & \makecell{49.87\\(0.10)} & \textbf{\makecell{18.96\\(0.35)}} & \makecell{48.39\\(0.13)}\\ \midrule
7 & $\mathrm{LSTM_{shuffle}}$ & \textbf{\makecell{87.45\\(0.27)}} & \textbf{\makecell{86.06\\(0.52)}} & \makecell{89.16\\(0.26)} & \textbf{\makecell{77.11\\(0.09)}} & \makecell{73.56\\(0.11)} & \textbf{\makecell{80.02\\(0.09)}} & \makecell{68.26\\(0.13)} & \makecell{48.21\\(1.60)} & \makecell{64.29\\(0.72)} & \makecell{69.49\\(0.20)} & \makecell{42.95\\(0.62)} & \makecell{69.01\\(0.20)} & \makecell{46.04\\(0.12)} & \makecell{16.21\\(0.11)} & \makecell{44.61\\(0.13)}\\ \midrule
8 & $\mathrm{LSTM_{set}}$ & \makecell{85.41\\(0.34)} & \makecell{84.60\\(0.37)} & \makecell{87.32\\(0.35)} & \makecell{76.52\\(0.08)} & \textbf{\makecell{73.83\\(0.12)}} & \makecell{78.98\\(0.09)} & \makecell{69.82\\(0.52)} & \makecell{53.82\\(0.47)} & \makecell{67.83\\(0.23)} & \makecell{69.42\\(0.59)} & \makecell{46.25\\(1.94)} & \makecell{68.64\\(0.43)} & \makecell{46.68\\(1.05)} & \makecell{18.85\\(0.23)} & \makecell{45.28\\(0.93)}\\ \midrule
9 & $\mathrm{TF_{set}}$ & \makecell{86.26\\(0.51)} & \makecell{85.10\\(0.55)} & \makecell{88.05\\(0.42)} & \makecell{59.74\\(32.58)} & \makecell{56.80\\(31.76)} & \makecell{61.70\\(34.01)} & \makecell{70.18\\(0.11)} & \makecell{53.70\\(0.85)} & \makecell{67.59\\(0.28)} & \textbf{\makecell{70.99\\(0.24)}} & \makecell{46.91\\(0.45)} & \makecell{70.03\\(0.26)} & \makecell{45.23\\(4.40)} & \makecell{18.71\\(1.69)} & \makecell{43.63\\(4.38)}\\ \midrule
10 & $\mathrm{FF_{sIoU}}$ & \makecell{87.25\\(0.07)} & \makecell{85.92\\(0.14)} & \textbf{\makecell{89.24\\(0.04)}} & \makecell{71.04\\(0.50)} & \makecell{57.11\\(1.47)} & \makecell{72.54\\(0.76)} & \makecell{64.13\\(1.45)} & \makecell{12.89\\(0.47)} & \makecell{61.47\\(5.96)} & \makecell{67.60\\(0.17)} & \makecell{20.73\\(0.40)} & \makecell{66.92\\(0.19)} & \makecell{44.23\\(0.35)} & \makecell{12.79\\(0.04)} & \makecell{42.45\\(0.33)}\\ \midrule
11 & $\mathrm{FF_{BCE,DC}}$ & \makecell{85.77\\(0.55)} & \makecell{84.11\\(0.58)} & \makecell{87.82\\(0.44)} & \makecell{70.63\\(0.90)} & \makecell{66.94\\(1.14)} & \makecell{72.19\\(0.69)} & \makecell{61.54\\(1.13)} & \makecell{44.23\\(1.63)} & \makecell{59.66\\(3.56)} & \makecell{70.75\\(0.24)} & \makecell{46.03\\(0.88)} & \textbf{\makecell{70.09\\(0.22)}} & \makecell{43.00\\(3.05)} & \makecell{14.87\\(1.25)} & \makecell{40.92\\(3.22)}\\ \midrule
12 & $\mathrm{FF_{sIoU,C}}$ & \makecell{85.89\\(0.08)} & \makecell{84.31\\(0.22)} & \makecell{88.01\\(0.10)} & \makecell{67.62\\(0.43)} & \makecell{50.66\\(1.95)} & \makecell{68.97\\(0.73)} & \makecell{63.12\\(0.38)} & \makecell{12.56\\(0.71)} & \makecell{63.69\\(0.39)} & \makecell{65.90\\(0.19)} & \makecell{19.88\\(0.31)} & \makecell{65.02\\(0.18)} & \makecell{42.78\\(0.25)} & \makecell{12.69\\(0.02)} & \makecell{40.69\\(0.25)}\\ \midrule
13 & $\mathrm{FF_{TD}}$ & \makecell{79.36\\(0.22)} & \makecell{78.43\\(0.32)} & \makecell{83.02\\(0.18)} & -- & -- & -- & -- & -- & -- & \makecell{64.73\\(0.37)} & \makecell{39.71\\(0.78)} & \makecell{64.56\\(0.33)} & \makecell{47.86\\(0.04)} & \makecell{18.34\\(0.07)} & \makecell{47.90\\(0.05)}\\
 \bottomrule
\end{tabular}}
\caption{\textbf{Set predictor comparison.} Results on VOC, COCO, NUS-WIDE, ADE20k and Recipe1M (test set), reported in terms of C-F1, O-F1 and I-F1. Models are trained 5 times using different random seeds. We report $mean$ ($std$) for each metric, model and dataset. Image representation backbone is fixed to ResNet-50. The models are ordered according to their average normalized O-F1 score computed over all five tested datasets. Note that $\mathrm{FF}_{\mathrm{TD}}$ is not considered to obtain the mean ranking, since it is not used for datasets including empty sets. Note that these results are not directly comparable with the ones from Table \ref{tab:sota_coco}.}
\label{tab:f1_metrics}
\vspace{-0.5cm}
\end{table*}

In this subsection, we aim to analyze the influence of the set predictor design. Table \ref{tab:f1_metrics} reports results for 13 set predictors built on top of a ResNet-50 image representation backbone. Each experiment was run with 5 different seeds (different from the one used for hyper-parameter selection). Models appear ranked following their average normalized O-F1 score over all datasets (O-F1 scores are normalized using the maximum O-F1 of the corresponding dataset). Interestingly, the simplest baseline $\mathrm{FF_{BCE}}$ consistently exhibits close to top performance across datasets. In particular, the baseline models shine in COCO, NUS-WIDE and ADE20k datasets. Auto-regressive models ($\mathrm{LSTM}$, $\mathrm{TF}$ and their shuffled versions) also rank favorably across most datasets, highlighting the potential role of modeling label co-occurrences. Moreover, their results suggests that, when there is an intrinsic label ordering in the dataset (e.g. NUS-WIDE and ADE20k preserve label order across dataset samples), exploiting it leads to increased performance. These findings raise the question of whether there exists an optimal per dataset label ordering to exploit, and support previous research along these lines \cite{Nam17setrnn,chen2018order}.

Taking a closer look at each dataset, the VOC top 3 performers across all metrics are $\mathrm{LSTM_{shuffle}}$, $\mathrm{FF_{sIoU}}$ and $\mathrm{TF_{shuffle}}$, all of which model label co-occurrences. Among the least performing models, we find $\mathrm{FF_{TD}}$ and all feed-forward models predicting set cardinality. These results suggest that modeling co-occurrences is beneficial, while exploiting label ordering and predicting set cardinality seems less impactful. 
The COCO top 3 performers are $\mathrm{LSTM_{shuffle}}$, $\mathrm{TF_{shuffle}}$ and $\mathrm{FF_{BCE}}$, which are closely followed by $\mathrm{LSTM}$ and $\mathrm{TF}$ (all within $1\%$ difference from the top performer). Again, among the least performing models, we find most of the feed-forward models which explicitly predict the set cardinality.\footnote{The low performance and high variance of $\mathrm{TF_{set}}$ is due to $2/5$ seeds that did not converge.}~
In this case, exploiting label ordering becomes more challenging as COCO labels only follow partial ordering.
For NUS-WIDE and ADE20k, top performers across all metrics also include auto-regressive models and $\mathrm{FF_{BCE}}$. As previously mentioned, those datasets have an pre-established label ordering, which favors $\mathrm{LSTM}$ and $\mathrm{TF}$ models vs their shuffled counterparts. By contrast to other datasets, endowing feed-forward models with cardinality prediction seems to achieve good performance in ADE20k. Finally, Recipe1M presents slightly different trends, with auto-regressive models among mid-performers. The top performers across all metrics are $\mathrm{FF_{BCE,C}}$ and $\mathrm{FF_{TD,C}}$, still advocating for the importance of cardinality prediction in feed-forward models when the dataset's dictionary size is large.

\begin{figure*}[t!]
\begin{minipage}[t]{0.33\linewidth}
    \centering
    \includegraphics[width=1.\columnwidth]{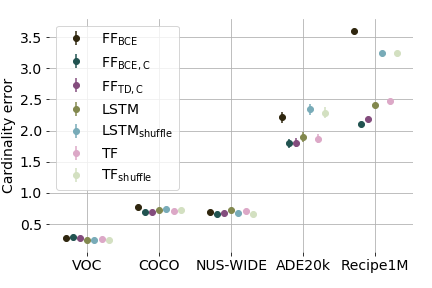}
    \caption{\textbf{Cardinality error per image (mean and 95\% confidence interval).} We compare the top 7 models from Table \ref{tab:f1_metrics}. }
    \label{fig:cardinality_error}
\end{minipage}\quad
\begin{minipage}[t]{0.66\linewidth}
    \begin{subfigure}[t]{0.5\textwidth}
        \includegraphics[width=\textwidth]{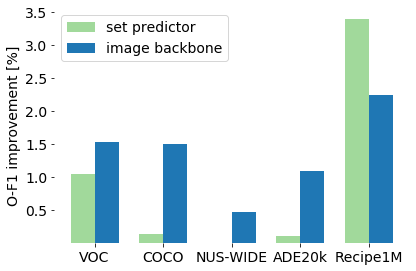}
        \caption{Comparison based on $\mathrm{FF_{BCE}}$}
        \label{fig:ff_bce_of1}
    \end{subfigure}\hfill
    \begin{subfigure}[t]{0.5\textwidth}
        \includegraphics[width=\textwidth]{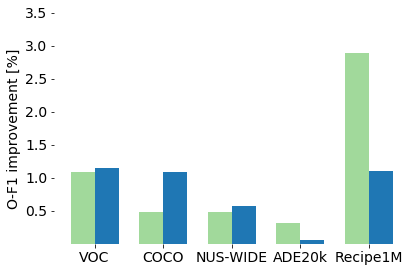}
         \caption{Comparison based on $\mathrm{LSTM}$}
        \label{fig:lstm_of1}
    \end{subfigure}\hfill
\caption{\textbf{O-F1 improvements when changing set predictor vs. image representation backbone.}}
\label{fig:backbone_vs_predictor}
\end{minipage}
\end{figure*}

Figure \ref{fig:cardinality_error} presents the cardinality prediction errors for the 7 best models of Table \ref{tab:f1_metrics}. As shown in the figure, the average cardinality error grows with the dataset's dictionary size, with VOC being the easiest dataset and Recipe1M the hardest. For Ade20k and Recipe1M datasets, feed-forward models explicitly trained to predict set cardinality (e. g. see $\mathrm{FF_{BCE}}$ vs. $\mathrm{FF_{BCE, C}}$) tend to be more accurate whereas for auto-regresive models, label shuffling leads to higher cardinality errors (e. g. see $\mathrm{LSTM_{shuffle}}$ vs $\mathrm{LSTM}$). Although an explicit cardinality prediction reduces the cardinality error, it does not always translate into better label prediction results, as highlighted in Table \ref{tab:f1_metrics}.

Moreover, it is worth noting that all models report significantly lower values for C-F1 than for the rest of the metrics in the case of NUS-WIDE, ADE20k and Recipe1M. These drops can be explained by the low class frequencies exhibited by a large number of classes (long-tail class distribution), which result in low average C-F1 when looking at low-frequency classes (with relatively large variance).  

\subsection{Impact of the image representation backbone}
In this subsection, we aim to assess the importance of the chosen image representation backbone. We compare the previously chosen ResNet-50 \cite{resnet} to ResNet-101 \cite{resnet} and ResNeXt-101-32x8d \cite{xie2016groups} for the 2 top ranked set predictors $\mathrm{FF_{BCE}}$ and $\mathrm{LSTM}$. Table \ref{tab:f1_backbone} reports the obtained results averaged across 5 different seeds (different from the one used for hyper-parameter selection in \textsc{Hyperband}). As shown in the table, for both set predictors, changing the image representation backbone from ResNet-50 to ResNet-101 or ResNeXt-101-32x8d leads to improvements in terms of all F1 scores. Improvements are especially notable for the $\mathrm{FF_{BCE}}$ set predictor, which gains over $1\%$ points in O-F1 in 4 out of 5 datasets, leading to a boost in performance when compared to the best models of Table \ref{tab:f1_metrics}. We hypothesize that feed forward predictors benefit more from enhanced single-label image classification backbones given their model design. Both single-label image classifiers and feed forward set predictors are composed of fully connected layers that are stacked on top of their respective convolutional backbones. As such, improvements in single-label image classification backbones translate into improvements in the multi-label scenario. Figures \ref{fig:ff_bce_of1} and \ref{fig:lstm_of1} highlight the O-F1 improvement that the best set predictor of Table \ref{tab:f1_metrics} achieves with respect to $\mathrm{FF_{BCE}}$ and $\mathrm{LSTM}$ respectively, for a fixed ResNet-50 image representation backbone. Similarly, the figure also displays the O-F1 improvement obtained when changing the image representation backbone of the same set predictors ($\mathrm{FF_{BCE}}$ and $\mathrm{LSTM}$). Interestingly, when it comes to $\mathrm{FF_{BCE}}$, exploiting better image representation backbones leads to higher improvements than enhancing the set predictor module while fixing the image representation backbone to ResNet-50. This observation also holds in 3 out 5 datasets in the case of the $\mathrm{LSTM}$ set predictor, which exhibits larger margins of improvement when changing the image representation backbone rather than the set predictor itself.

\begin{table*}[t!]
\centering
\resizebox{\textwidth}{!}{
\begin{tabular}{@{}llccc|ccc|ccc|ccc|ccc@{}}
\toprule
 & & \multicolumn{3}{c}{\textbf{VOC}}        & \multicolumn{3}{c}{\textbf{COCO}}          & \multicolumn{3}{c}{\textbf{NUS-WIDE}}      & \multicolumn{3}{c}{\textbf{ADE20k}}        & \multicolumn{3}{c}{\textbf{Recipe1M}}   \\
\textbf{Model} &\textbf{Backbone}                       & \textbf{O-F1}      & \textbf{C-F1}     & \textbf{I-F1}     & \textbf{O-F1}      & \textbf{C-F1}     & \textbf{I-F1}     & \textbf{O-F1}      & \textbf{C-F1}     & \textbf{I-F1}      & \textbf{O-F1}      & \textbf{C-F1}     & \textbf{I-F1}    & \textbf{O-F1}      & \textbf{C-F1}     & \textbf{I-F1}    \\ \specialrule{1pt}{1.25\jot}{1.25\jot}
 
  & ResNet-50 &\makecell{86.40\\(0.19)}&\makecell{85.16\\(0.20)}&\makecell{88.23\\(0.15)}&\makecell{76.98\\(0.06)}&\makecell{73.49\\(0.14)}&\makecell{79.35\\(0.07)}&\makecell{71.34\\(0.06)}&\makecell{54.94\\(0.48)}&\makecell{69.19\\(0.12)}&\makecell{70.89\\(0.34)}&\makecell{47.87\\(1.28)}&\makecell{69.43\\(0.30)}&\makecell{46.83\\(0.07)}&\makecell{18.38\\(0.17)}&\makecell{43.63\\(0.07)}\\ \cmidrule(lr){2-17}$\mathrm{FF_{BCE}}$ & ResNet-101 &\makecell{87.64\\(0.14)}&\makecell{86.57\\(0.23)}&\makecell{89.34\\(0.12)}&\makecell{77.67\\(0.07)}&\makecell{74.78\\(0.11)}&\makecell{79.80\\(0.10)}&\textbf{\makecell{71.81\\(0.04)}} &\makecell{57.82\\(0.07)}&\textbf{\makecell{69.61\\(0.09)}} &\makecell{71.81\\(0.23)}&\textbf{\makecell{49.92\\(0.60)}} &\makecell{70.81\\(0.22)}&\makecell{47.90\\(0.12)}&\makecell{18.59\\(0.11)}&\makecell{44.78\\(0.12)}\\ \cmidrule(lr){2-17}  & ResNeXt-101-32x8d &\textbf{\makecell{87.93\\(0.20)}} &\textbf{\makecell{87.60\\(0.13)}} &\textbf{\makecell{90.05\\(0.18)}} &\textbf{\makecell{78.48\\(0.09)}} &\textbf{\makecell{75.62\\(0.19)}} &\textbf{\makecell{80.88\\(0.12)}} &\makecell{71.72\\(0.02)}&\textbf{\makecell{57.95\\(0.10)}} &\makecell{69.34\\(0.03)}&\textbf{\makecell{71.98\\(0.21)}} &\makecell{46.25\\(0.44)}&\textbf{\makecell{70.87\\(0.24)}} &\textbf{\makecell{49.07\\(0.10)}} &\textbf{\makecell{19.13\\(0.02)}} &\textbf{\makecell{46.12\\(0.12)}} \\ \specialrule{1pt}{1.25\jot}{1.25\jot} & ResNet-50 &\makecell{86.36\\(0.18)}&\makecell{85.00\\(0.20)}&\makecell{88.20\\(0.09)}&\makecell{76.63\\(0.08)}&\makecell{72.98\\(0.07)}&\makecell{79.45\\(0.09)}&\makecell{70.85\\(0.07)}&\makecell{54.15\\(0.16)}&\makecell{69.43\\(0.04)}&\makecell{70.68\\(0.23)}&\textbf{\makecell{48.73\\(1.40)}} &\makecell{69.97\\(0.23)}&\makecell{47.33\\(0.05)}&\makecell{17.55\\(0.05)}&\makecell{46.12\\(0.06)}\\ \cmidrule(lr){2-17}$\mathrm{LSTM}$ & ResNet-101 &\makecell{87.29\\(0.32)}&\makecell{85.77\\(0.43)}&\makecell{89.01\\(0.31)}&\makecell{77.62\\(0.06)}&\makecell{74.39\\(0.12)}&\makecell{80.27\\(0.07)}&\textbf{\makecell{71.43\\(0.03)}} &\textbf{\makecell{55.76\\(0.24)}} &\textbf{\makecell{70.08\\(0.04)}} &\textbf{\makecell{70.73\\(0.11)}} &\makecell{48.13\\(0.41)}&\textbf{\makecell{70.06\\(0.20)}} &\makecell{48.03\\(0.07)}&\makecell{17.87\\(0.05)}&\makecell{46.84\\(0.08)}\\ \cmidrule(lr){2-17}  & ResNeXt-101-32x8d &\textbf{\makecell{87.51\\(0.16)}} &\textbf{\makecell{86.53\\(0.16)}} &\textbf{\makecell{89.40\\(0.18)}} &\textbf{\makecell{77.71\\(0.07)}} &\textbf{\makecell{74.74\\(0.12)}} &\textbf{\makecell{80.34\\(0.08)}} &\makecell{70.86\\(0.16)}&\makecell{54.26\\(0.36)}&\makecell{69.58\\(0.23)}&\makecell{70.00\\(0.20)}&\makecell{47.84\\(1.14)}&\makecell{69.23\\(0.29)}&\textbf{\makecell{48.43\\(0.06)}} &\textbf{\makecell{18.18\\(0.15)}} &\textbf{\makecell{47.23\\(0.08)}} \\
 \bottomrule
\end{tabular}}
\caption{\textbf{Image representation backbone comparison.} Results on VOC, COCO, NUS-WIDE, ADE20k and Recipe1M (test set) reported in terms of C-F1, O-F1 and I-F1. Models are trained 5 times using different random seeds. We report $mean$ ($std$) for each metric, set predictor, image representation backbone and dataset. Note that these results are not directly comparable with the ones from Table \ref{tab:sota_coco}.}
\label{tab:f1_backbone}
\vspace{-0.5cm}
\end{table*}

\subsection{Summary of observations}

On the one hand, our experiments have shown that image-to-set prediction backbones follow the trends of single label classification, i.e. ResNext-101 improving results over both ResNet-50 and ResNet-101. Moreover, our results suggest that, in many cases, enhancing the image representation backbone can provide larger performance boosts than enhancing the set predictors. Therefore, when designing new image-to-set prediction models, we recommend taking advantage of the top performing single-label classification architectures.

On the other hand, when it comes to set predictors, auto-regressive models which explicitly leverage label co-occurrences seem to achieve slightly better overall results. In that case, exploiting dataset order is beneficial for datasets that consistently present labels following the same order. Including cardinality prediction in feed forward models seems to be relevant in cases where the dictionary size is very large (e.g. in Recipe1M). One must not disregard the high performance achieved by the simple yet effective baseline model, namely $\mathrm{FF_{BCE}}$, which has the potential to achieve the best performing results when given the best image representation backbone.

Finally, it is important to note that we experienced a significant performance boost (in all methods) when employing an automatic hyper-parameter strategy. In this case, it is important to allocate the same search budget to all models and baselines to ensure a fair comparison and robust conclusions on the proposed approach.

\section{Conclusion}

In this paper, we have described important reproducibility challenges in the image-to-set prediction literature that impede proper comparisons among published methods and hinder the research progress. To alleviate this issue, we equipped the community with a benchmark suite composed of a unified code-base, predefined splits for 5 datasets of increasing complexity and a common evaluation protocol to assess the impact of current design choices and future innovations. Together with the benchmark suite release, we performed an in-depth analysis of the key components of current image-to-set prediction models, namely the choice of image representation backbone as well as the set predictor design. In total, we compared 3 different image representation backbones and 13 different families of set predictors. To ensure fair and robust comparisons among methods, we used the \textsc{Hyperband} algorithm with a fixed budged of tested configurations ($410$ hyperparameter configurations evaluated per model) and reported results averaged over 5 different seeds (different than that used for tuning).

Looking forward, we expect that the release of the benchmark suite and the performed analysis will accelerate research in the image-to-set prediction domain, by ensuring firm steps and robust conclusions out of future contributions. As general guidance, when introducing new set prediction approaches, we suggest:
\begin{compactitem}
   \item[--] Initializing the image-to-set prediction backbone with ImageNet pre-trained models.
   \item[--] Performing hyper-parameter search of the newly introduced methods by fixing a limited budget of configurations to be tested (and applying the same budget to baselines).
   \item[--] Using the suggested validation set to perform architectural and optimization hyper-parameter search, including early-stopping.
   \item[--] Checking the test set performance only once, after finalizing the hyper-parameter tuning.
   \item[--] Reporting results on the test set with multiple image representation backbones and training seeds.
\end{compactitem}
Together with the benchmark suite, we release a subset of the top performing models per dataset. These pre-trained models are meant to be used for future comparisons and, potentially, transfer learning. 

{\small
\bibliographystyle{ieee_fullname}
\bibliography{image2set_biblio}

\begin{thebibliography}{10}\itemsep=-1pt

\bibitem{AntonucciCMG13}
Alessandro Antonucci, Giorgio Corani, Denis~Deratani Mau{\'{a}}, and Sandra
  Gabaglio.
\newblock An ensemble of bayesian networks for multilabel classification.
\newblock In {\em IJCAI}, 2013.

\bibitem{Bi2013}
Wei Bi and James~T. Kwok.
\newblock Efficient multi-label classification with many labels.
\newblock In {\em ICML}, 2013.

\bibitem{Bouthillier19a}
Xavier Bouthillier, C{\'e}sar Laurent, and Pascal Vincent.
\newblock Unreproducible research is reproducible.
\newblock In Kamalika Chaudhuri and Ruslan Salakhutdinov, editors, {\em
  Proceedings of the 36th International Conference on Machine Learning},
  volume~97 of {\em Proceedings of Machine Learning Research}, pages 725--734,
  Long Beach, California, USA, 09--15 Jun 2019. PMLR.

\bibitem{chatfield2014return}
Ken Chatfield, Karen Simonyan, Andrea Vedaldi, and Andrew Zisserman.
\newblock Return of the devil in the details: Delving deep into convolutional
  nets.
\newblock {\em BMVC}, 2014.

\bibitem{chen2016deep}
Jing-Jing Chen and Chong-Wah Ngo.
\newblock Deep-based ingredient recognition for cooking recipe retrieval.
\newblock In {\em ACM Multimedia}, 2016.

\bibitem{chen2018order}
Shang-Fu Chen, Yi-Chen Chen, Chih-Kuan Yeh, and Yu-Chiang~Frank Wang.
\newblock Order-free rnn with visual attention for multi-label classification.
\newblock In {\em AAAI}, 2018.

\bibitem{chen2018recurrent}
Tianshui Chen, Zhouxia Wang, Guanbin Li, and Liang Lin.
\newblock Recurrent attentional reinforcement learning for multi-label image
  recognition.
\newblock In {\em AAAI}, 2018.

\bibitem{chen2019multi}
Zhao-Min Chen, Xiu-Shen Wei, Peng Wang, and Yanwen Guo.
\newblock Multi-label image recognition with graph convolutional networks.
\newblock In {\em CVPR}, 2019.

\bibitem{chua2009nus}
Tat-Seng Chua, Jinhui Tang, Richang Hong, Haojie Li, Zhiping Luo, and Yantao
  Zheng.
\newblock Nus-wide: a real-world web image database from national university of
  singapore.
\newblock In {\em ACM Conference on Image and Video Retrieval}, 2009.

\bibitem{cityscapes}
Marius Cordts, Mohamed Omran, Sebastian Ramos, Timo Rehfeld, Markus Enzweiler,
  Rodrigo Benenson, Uwe Franke, Stefan Roth, and Bernt Schiele.
\newblock The cityscapes dataset for semantic urban scene understanding.
\newblock In {\em The IEEE Conference on Computer Vision and Pattern
  Recognition (CVPR)}, June 2016.

\bibitem{Dembczynski2010pcc}
Krzysztof Dembczy\'{n}ski, Weiwei Cheng, and Eyke H\"{u}llermeier.
\newblock Bayes optimal multilabel classification via probabilistic classifier
  chains.
\newblock In {\em ICML}, 2010.

\bibitem{Drozdzal2016}
Michal Drozdzal, Eugene Vorontsov, Gabriel Chartrand, Samuel Kadoury, and Chris
  Pal.
\newblock The importance of skip connections in biomedical image segmentation.
\newblock In Gustavo Carneiro, Diana Mateus, Lo{\"i}c Peter, Andrew Bradley,
  Jo{\~a}o Manuel R.~S. Tavares, Vasileios Belagiannis, Jo{\~a}o~Paulo Papa,
  Jacinto~C. Nascimento, Marco Loog, Zhi Lu, Jaime~S. Cardoso, and Julien
  Cornebise, editors, {\em Deep Learning and Data Labeling for Medical
  Applications}, pages 179--187, Cham, 2016. Springer International Publishing.

\bibitem{everingham2010pascal}
Mark Everingham, Luc Van~Gool, Christopher~KI Williams, John Winn, and Andrew
  Zisserman.
\newblock The pascal visual object classes (voc) challenge.
\newblock {\em IJCV}, 2010.

\bibitem{ge2018multi}
Weifeng Ge, Sibei Yang, and Yizhou Yu.
\newblock Multi-evidence filtering and fusion for multi-label classification,
  object detection and semantic segmentation based on weakly supervised
  learning.
\newblock In {\em CVPR}, 2018.

\bibitem{GongJLTI13}
Yunchao Gong, Yangqing Jia, Thomas Leung, Alexander Toshev, and Sergey Ioffe.
\newblock Deep convolutional ranking for multilabel image annotation.
\newblock {\em CoRR}, abs/1312.4894, 2013.

\bibitem{guo2019visual}
Hao Guo, Kang Zheng, Xiaochuan Fan, Hongkai Yu, and Song Wang.
\newblock Visual attention consistency under image transforms for multi-label
  image classification.
\newblock In {\em CVPR}, 2019.

\bibitem{maskrcnn}
Kaiming He, Georgia Gkioxari, Piotr Doll{\'a}r, and Ross Girshick.
\newblock Mask r-cnn.
\newblock In {\em ICCV}, 2017.

\bibitem{resnet}
Kaiming He, Xiangyu Zhang, Shaoqing Ren, and Jian Sun.
\newblock Deep residual learning for image recognition.
\newblock In {\em CVPR}, 2016.

\bibitem{Henderson2017}
Peter Henderson, Riashat Islam, Philip Bachman, Joelle Pineau, Doina Precup,
  and David Meger.
\newblock Deep reinforcement learning that matters.
\newblock {\em CoRR}, abs/1709.06560, 2017.

\bibitem{lstm}
Sepp Hochreiter and J{\"u}rgen Schmidhuber.
\newblock Long short-term memory.
\newblock {\em Neural computation}, 1997.

\bibitem{NIPS2009_3824}
Daniel~J Hsu, Sham~M Kakade, John Langford, and Tong Zhang.
\newblock Multi-label prediction via compressed sensing.
\newblock In Y. Bengio, D. Schuurmans, J.~D. Lafferty, C.~K.~I. Williams, and
  A. Culotta, editors, {\em NeurIPS}. Curran Associates, Inc., 2009.

\bibitem{jamieson2016non}
Kevin Jamieson and Ameet Talwalkar.
\newblock Non-stochastic best arm identification and hyperparameter
  optimization.
\newblock In {\em AISTATS}, pages 240--248, 2016.

\bibitem{jegou2017one}
Simon J{\'e}gou, Michal Drozdzal, David Vazquez, Adriana Romero, and Yoshua
  Bengio.
\newblock The one hundred layers tiramisu: Fully convolutional densenets for
  semantic segmentation.
\newblock In {\em CVPR-W}, 2017.

\bibitem{NIPS2012_4591}
Ashish Kapoor, Raajay Viswanathan, and Prateek Jain.
\newblock Multilabel classification using bayesian compressed sensing.
\newblock In F. Pereira, C.~J.~C. Burges, L. Bottou, and K.~Q. Weinberger,
  editors, {\em NeurIPS}. Curran Associates, Inc., 2012.

\bibitem{KingmaB14}
Diederik~P. Kingma and Jimmy Ba.
\newblock Adam: {A} method for stochastic optimization.
\newblock {\em CoRR}, abs/1412.6980, 2014.

\bibitem{Kurach2018}
Karol Kurach, Mario Lucic, Xiaohua Zhai, Marcin Michalski, and Sylvain Gelly.
\newblock The {GAN} landscape: Losses, architectures, regularization, and
  normalization.
\newblock {\em CoRR}, abs/1807.04720, 2018.

\bibitem{li2018hyperband}
Lisha Li, Kevin Jamieson, Giulia DeSalvo, Afshin Rostamizadeh, and Ameet
  Talwalkar.
\newblock Hyperband: A novel bandit-based approach to hyperparameter
  optimization.
\newblock {\em JMLR}, 18(185):1--52, 2018.

\bibitem{li2018attentive}
Liang Li, Shuhui Wang, Shuqiang Jiang, and Qingming Huang.
\newblock Attentive recurrent neural network for weak-supervised multi-label
  image classification.
\newblock In {\em ACM Multimedia}, 2018.

\bibitem{li2017improving}
Yuncheng Li, Yale Song, and Jiebo Luo.
\newblock Improving pairwise ranking for multi-label image classification.
\newblock In {\em CVPR}, 2017.

\bibitem{lin2014microsoft}
Tsung-Yi Lin, Michael Maire, Serge Belongie, James Hays, Pietro Perona, Deva
  Ramanan, Piotr Doll{\'a}r, and C~Lawrence Zitnick.
\newblock Microsoft coco: Common objects in context.
\newblock In {\em ECCV}, 2014.

\bibitem{Lin2014MCV}
Zijia Lin, Guiguang Ding, Mingqing Hu, and Jianmin Wang.
\newblock Multi-label classification via feature-aware implicit label space
  encoding.
\newblock In {\em ICML}, 2014.

\bibitem{liu2017semantic}
Feng Liu, Tao Xiang, Timothy~M Hospedales, Wankou Yang, and Changyin Sun.
\newblock Semantic regularisation for recurrent image annotation.
\newblock In {\em CVPR}, 2017.

\bibitem{liu2019decoupling}
Luchen Liu, Sheng Guo, Weilin Huang, and Matthew~R Scott.
\newblock Decoupling category-wise independence and relevance with
  self-attention for multi-label image classification.
\newblock In {\em ICASSP}, 2019.

\bibitem{Liu2015LargeMM}
Weiwei Liu and Ivor~W. Tsang.
\newblock Large margin metric learning for multi-label prediction.
\newblock In {\em AAAI}, 2015.

\bibitem{liu2018multi}
Yongcheng Liu, Lu Sheng, Jing Shao, Junjie Yan, Shiming Xiang, and Chunhong
  Pan.
\newblock Multi-label image classification via knowledge distillation from
  weakly-supervised detection.
\newblock In {\em ACM Multimedia}, 2018.

\bibitem{Locatello2018}
Francesco Locatello, Stefan Bauer, Mario Lucic, Sylvain Gelly, Bernhard
  Sch{\"{o}}lkopf, and Olivier Bachem.
\newblock Challenging common assumptions in the unsupervised learning of
  disentangled representations.
\newblock {\em CoRR}, abs/1811.12359, 2018.

\bibitem{long2015fully}
Jonathan Long, Evan Shelhamer, and Trevor Darrell.
\newblock Fully convolutional networks for semantic segmentation.
\newblock In {\em CVPR}, 2015.

\bibitem{lu2017knowing}
Jiasen Lu, Caiming Xiong, Devi Parikh, and Richard Socher.
\newblock Knowing when to look: Adaptive attention via a visual sentinel for
  image captioning.
\newblock In {\em CVPR}, 2017.

\bibitem{Lucic2018}
Mario Lucic, Karol Kurach, Marcin Michalski, Sylvain Gelly, and Olivier
  Bousquet.
\newblock Are gans created equal? a large-scale study.
\newblock In S. Bengio, H. Wallach, H. Larochelle, K. Grauman, N. Cesa-Bianchi,
  and R. Garnett, editors, {\em Advances in Neural Information Processing
  Systems 31}, pages 700--709. Curran Associates, Inc., 2018.

\bibitem{luo2019visual}
Yan Luo, Ming Jiang, and Qi Zhao.
\newblock Visual attention in multi-label image classification.
\newblock In {\em CVPR-W}, 2019.

\bibitem{lyu2019attend}
Fan Lyu, Qi Wu, Fuyuan Hu, Qingyao Wu, and Mingkui Tan.
\newblock Attend and imagine: Multi-label image classification with visual
  attention and recurrent neural networks.
\newblock {\em IEEE Transactions on Multimedia}, 2019.

\bibitem{Mahajan18hashtags}
Dhruv Mahajan, Ross~B. Girshick, Vignesh Ramanathan, Kaiming He, Manohar
  Paluri, Yixuan Li, Ashwin Bharambe, and Laurens van~der Maaten.
\newblock Exploring the limits of weakly supervised pretraining.
\newblock {\em CoRR}, abs/1805.00932, 2018.

\bibitem{Melis2017}
G{\'{a}}bor Melis, Chris Dyer, and Phil Blunsom.
\newblock On the state of the art of evaluation in neural language models.
\newblock {\em CoRR}, abs/1707.05589, 2017.

\bibitem{Nam2014}
Jinseok Nam, Jungi Kim, Eneldo~Loza Menc\'{\i}a, Iryna Gurevych, and Johannes
  F\"{u}rnkranz.
\newblock Large-scale multi-label text classification --- revisiting neural
  networks.
\newblock In {\em ECMLPKDD}, 2014.

\bibitem{Nam17setrnn}
Jinseok Nam, Eneldo Loza~Menc\'{\i}a, Hyunwoo~J Kim, and Johannes
  F\"{u}rnkranz.
\newblock Maximizing subset accuracy with recurrent neural networks in
  multi-label classification.
\newblock In {\em NeurIPS}, 2017.

\bibitem{nauata2018structured}
Nelson Nauata, Hexiang Hu, Guang-Tong Zhou, Zhiwei Deng, Zicheng Liao, and Greg
  Mori.
\newblock Structured label inference for visual understanding.
\newblock {\em TPAMI}, 2019.

\bibitem{Oliver2018}
Avital Oliver, Augustus Odena, Colin Raffel, Ekin~D. Cubuk, and Ian~J.
  Goodfellow.
\newblock Realistic evaluation of deep semi-supervised learning algorithms.
\newblock {\em CoRR}, abs/1804.09170, 2018.

\bibitem{paszke2017automatic}
Adam Paszke, Sam Gross, Soumith Chintala, Gregory Chanan, Edward Yang, Zachary
  DeVito, Zeming Lin, Alban Desmaison, Luca Antiga, and Adam Lerer.
\newblock Automatic differentiation in pytorch.
\newblock In {\em NeurIPS-W}, 2017.

\bibitem{yolo}
Joseph Redmon, Santosh Divvala, Ross Girshick, and Ali Farhadi.
\newblock You only look once: Unified, real-time object detection.
\newblock In {\em CVPR}, 2016.

\bibitem{faster-rcnn}
Shaoqing Ren, Kaiming He, Ross Girshick, and Jian Sun.
\newblock Faster {R-CNN:} towards real-time object detection with region
  proposal networks.
\newblock In {\em NeurIPS}, 2015.

\bibitem{deepsetnet}
S~Hamid Rezatofighi, Anton Milan, Ehsan Abbasnejad, Anthony Dick, Ian Reid,
  et~al.
\newblock Deepsetnet: Predicting sets with deep neural networks.
\newblock In {\em ICCV}, 2017.

\bibitem{rezatofighi2017joint}
S~Hamid Rezatofighi, Anton Milan, Qinfeng Shi, Anthony Dick, and Ian Reid.
\newblock Joint learning of set cardinality and state distribution.
\newblock {\em AAAI}, 2018.

\bibitem{imagenet}
Olga Russakovsky, Jia Deng, Hao Su, Jonathan Krause, Sanjeev Satheesh, Sean Ma,
  Zhiheng Huang, Andrej Karpathy, Aditya Khosla, Michael Bernstein,
  Alexander~C. Berg, and Li Fei-Fei.
\newblock {ImageNet Large Scale Visual Recognition Challenge}.
\newblock {\em IJCV}, 2015.

\bibitem{inversecooking}
Amaia Salvador, Michal Drozdzal, Xavier {Gir{\'{o}} i Nieto}, and Adriana
  Romero.
\newblock Inverse cooking: Recipe generation from food images.
\newblock In {\em CVPR}, 2019.

\bibitem{salvador2017learning}
Amaia Salvador, Nicholas Hynes, Yusuf Aytar, Javier Marin, Ferda Ofli, Ingmar
  Weber, and Antonio Torralba.
\newblock Learning cross-modal embeddings for cooking recipes and food images.
\newblock In {\em CVPR}, 2017.

\bibitem{ShuLXT15}
Xin Shu, Darong Lai, Huanliang Xu, and Liang Tao.
\newblock Learning shared subspace for multi-label dimensionality reduction via
  dependence maximization.
\newblock {\em Neurocomputing}, 2015.

\bibitem{vgg}
Karen Simonyan and Andrew Zisserman.
\newblock Very deep convolutional networks for large-scale image recognition.
\newblock In {\em ICLR}, 2015.

\bibitem{SongMVK15}
Yale Song, Daniel McDuff, Deepak Vasisht, and Ashish Kapoor.
\newblock Exploiting sparsity and co-occurrence structure for action unit
  recognition.
\newblock In {\em {FG}}. {IEEE} Computer Society, 2015.

\bibitem{Tsoumakas10powerset}
Grigorios Tsoumakas and Ioannis Vlahavas.
\newblock Random k-labelsets: An ensemble method for multilabel classification.
\newblock In Joost~N. Kok, Jacek Koronacki, Raomon Lopez~de Mantaras, Stan
  Matwin, Dunja Mladeni{\v{c}}, and Andrzej Skowron, editors, {\em ECML}, 2007.

\bibitem{transformer}
Ashish Vaswani, Noam Shazeer, Niki Parmar, Jakob Uszkoreit, Llion Jones,
  Aidan~N Gomez, {\L}ukasz Kaiser, and Illia Polosukhin.
\newblock Attention is all you need.
\newblock In {\em NeurIPS}, 2017.

\bibitem{WangYMHHX16cnnrnn}
Jiang Wang, Yi Yang, Junhua Mao, Zhiheng Huang, Chang Huang, and Wei Xu.
\newblock {CNN-RNN:} {A} unified framework for multi-label image
  classification.
\newblock In {\em CVPR}, 2016.

\bibitem{wang2017multi}
Zhouxia Wang, Tianshui Chen, Guanbin Li, Ruijia Xu, and Liang Lin.
\newblock Multi-label image recognition by recurrently discovering attentional
  regions.
\newblock In {\em ICCV}, 2017.

\bibitem{WeiXHNDZY14}
Yunchao Wei, Wei Xia, Junshi Huang, Bingbing Ni, Jian Dong, Yao Zhao, and
  Shuicheng Yan.
\newblock {CNN:} single-label to multi-label.
\newblock {\em CoRR}, abs/1406.5726, 2014.

\bibitem{wei2016hcp}
Yunchao Wei, Wei Xia, Min Lin, Junshi Huang, Bingbing Ni, Jian Dong, Yao Zhao,
  and Shuicheng Yan.
\newblock Hcp: A flexible cnn framework for multi-label image classification.
\newblock {\em TPAMI}, 2016.

\bibitem{NIPS2018_7820}
Sean Welleck, Zixin Yao, Yu Gai, Jialin Mao, Zheng Zhang, and Kyunghyun Cho.
\newblock Loss functions for multiset prediction.
\newblock In S. Bengio, H. Wallach, H. Larochelle, K. Grauman, N. Cesa-Bianchi,
  and R. Garnett, editors, {\em Advances in Neural Information Processing
  Systems 31}, pages 5783--5792. Curran Associates, Inc., 2018.

\bibitem{Weston2011WSU}
Jason Weston, Samy Bengio, and Nicolas Usunier.
\newblock Wsabie: Scaling up to large vocabulary image annotation.
\newblock In {\em IJCAI}, 2011.

\bibitem{WuJLGL18}
Baoyuan Wu, Fan Jia, Wei Liu, Bernard Ghanem, and Siwei Lyu.
\newblock Multi-label learning with missing labels using mixed dependency
  graphs.
\newblock {\em IJCV}, 2018.

\bibitem{xie2016groups}
Saining Xie, Ross Girshick, Piotr Doll\'{a}r, Zhuowen Tu, and Kaiming He.
\newblock Aggregated residual transformations for deep neural networks.
\newblock In {\em {CVPR}}, 2017.

\bibitem{yang2016exploit}
Hao Yang, Joey Tianyi~Zhou, Yu Zhang, Bin-Bin Gao, Jianxin Wu, and Jianfei Cai.
\newblock Exploit bounding box annotations for multi-label object recognition.
\newblock In {\em CVPR}, 2016.

\bibitem{YehWKW17}
Chih{-}Kuan Yeh, Wei{-}Chieh Wu, Wei{-}Jen Ko, and Yu{-}Chiang~Frank Wang.
\newblock Learning deep latent spaces for multi-label classification.
\newblock {\em CoRR}, abs/1707.00418, 2017.

\bibitem{zhang2018multilabel}
Junjie Zhang, Qi Wu, Chunhua Shen, Jian Zhang, and Jianfeng Lu.
\newblock Multilabel image classification with regional latent semantic
  dependencies.
\newblock {\em IEEE Transactions on Multimedia}, 2018.

\bibitem{Zhang2007}
Min-Ling Zhang and Zhi-Hua Zhou.
\newblock Ml-knn: A lazy learning approach to multi-label learning.
\newblock {\em Pattern Recogn.}, 40(7), July 2007.

\bibitem{Zhao2015}
Feipeng Zhao and Yuhong Guo.
\newblock Semi-supervised multi-label learning with incomplete labels.
\newblock In {\em IJCAI}, 2015.

\bibitem{zhou2017scene}
Bolei Zhou, Hang Zhao, Xavier Puig, Sanja Fidler, Adela Barriuso, and Antonio
  Torralba.
\newblock Scene parsing through ade20k dataset.
\newblock In {\em CVPR}, 2017.

\bibitem{zhu2017learning}
Feng Zhu, Hongsheng Li, Wanli Ouyang, Nenghai Yu, and Xiaogang Wang.
\newblock Learning spatial regularization with image-level supervisions for
  multi-label image classification.
\newblock In {\em CVPR}, 2017.

\end{thebibliography}
}

\clearpage
\appendix
\title{Supplementary material\vspace{-3em}}
\date{}
\author{}
\maketitle

We start the supplementary material by providing a short description of the datasets used (Section \ref{dataset}). Then, in Section \ref{implementation}, we provide implementation details, including data pre-processing, hyperparamenter values considered for the \textsc{Hyperband} tunning, as well as values of the selected hyperparamenters. 

\section{Datasets details}
\label{dataset}

We train and evaluate our models on five different image datasets, which provide multi-label annotations. 

\noindent \textbf{Pascal VOC 2007~\cite{everingham2010pascal}} is a popular benchmark for image classification, object detection and segmentation tasks. It is composed of $9\,963$ images containing objects from $20$ distinct categories. Images are divided in $2\,501$, $2\,510$ and $4\,952$ for train, validation and test splits, respectively. We train with $90\%$ of the \emph{trainval} images, keeping $10\%$ for validation. Models are evaluated on the test set, for which annotations have been released.  

\noindent \textbf{MS COCO 2014~\cite{lin2014microsoft}} is a popular benchmark for object detection and segmentation on natural images, containing annotations for objects of $80$ different categories. It is composed of $82\,783$ images for training and $40\,504$ for validation. Since evaluation on the test set can only be done through the benchmark server, which currently does not support the set prediction task, we use $10\%$ of the training set for validation, and evaluate on the full validation set. Note that in our experiments we include images with no annotations as \emph{empty sets}. 

\noindent \textbf{NUS-WIDE~\cite{chua2009nus}} is a web image database composed of $161\,789$ images for training and $107\,859$ for testing, annotated with $81$ unique tags collected from Flickr. While VOC and MS COCO are annotated with visually grounded object tags (e.g. \emph{dog}, \emph{train} or \emph{person}), NUS-WIDE includes a wider variety of tags referring to activities (e.g. \emph{wedding}, \emph{soccer}), scenes (e.g. \emph{snow}, \emph{airport}) and objects (e.g. \emph{car}, \emph{computer}, \emph{dog}). As in COCO, this dataset includes images with \emph{empty sets} annotations.

\noindent \textbf{ADE20k~\cite{zhou2017scene}} is a scene parsing dataset, containing $20\,210$ training, $2\,000$ validation samples, annotated with a dictionary of $150$ labels. Since the test set server evaluation is not suited for image to set prediction, we use validation set as a test set and separate a new validation set from the training set. As a result we obtain $18\,176$, $2\,020$ and $2\,000$ images for train, validation and test splits, respectively.

\noindent \textbf{Recipe1M~\cite{salvador2017learning}} composed of $1\,029\,720$ recipes scraped from cooking websites. The dataset is split in $720\,639$ training, $155\,036$ validation and $ 154\,045$ test samples, each containing a cooking recipe (from which we only use ingredients) and (optionally) images. In our experiments, we use only those samples containing images. Following \cite{inversecooking}, we pre-process the ingredient dictionary by (1) removing plurals, (2) clustering together ingredients that share the first or last two words, (3) merge ingredients sharing the first or last word, (4) removing infrequent ingredients (appearing less than $10$ times), and (5) remove recipes with less than $2$ ingredients. This procedure results in $1\,486$ unique ingredients and  $252\,547$ training, $54\,255$ validation and $54\,506$ test samples. To speed up the training, we use $5\,000$ randomly chosen validation images.
\setcounter{subfigure}{0}

\subsection{Order in dataset labels}
\label{app:label_order}
\begin{figure}
    \centering
    \includegraphics[width=\columnwidth]{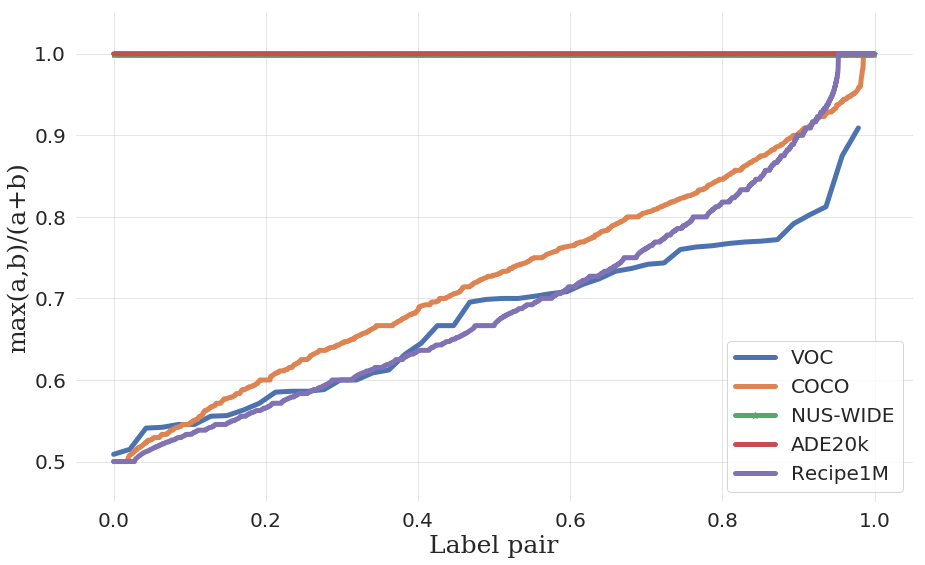}
    \caption{\textbf{Order in label pairs.}}
    \label{fig:supp:order_pairs}
\end{figure}

In this Subsection, we demonstrate that some datasets come with a preexisting label order while other do not. Figure \ref{fig:supp:order_pairs} depicts the order in label pairs for each dataset. The x-axis is normalized for each dataset. For each label pair ($A$, $B$), we compute the number of times that one label precedes the other. Then, we compute order $O = max(a,b)/(a+b)$, where $a$ accounts for the number of times that $A$ precedes $B$ in the set (and vice versa for $b$). A value of $O = 0.5$ indicates no order (i.e. $A$ precedes $B$ as often as $B$ precedes $A$), and a value of $1.0$ indicates total order ($A$ always precedes $B$, or vice versa). In the case of NUS-WIDE and ADE20k, labels always appear in the same order for all datapoints. For VOC, COCO and Recipe1M, while the plot reveals some degree of order for all label pairs (all values are above $0.5$), most values are below $1.0$, indicating that label order is not consistent across samples.

\section{Implementation details}
\label{implementation}

\subsection{Data pre-processing}
We resize all images to $448$ pixels in their shortest side, preserving aspect ratio, and take random crops of $448\times448$ for training. We randomly flip ($p = 0.5$), translate (within a range of $\pm 10\%$ of the image size on each axis) and rotate images ($\pm 10$\textdegree) for data augmentation during training. All models are trained with the Adam optimizer \cite{KingmaB14} for a maximum of $200$ epochs, or until early-stopping criterion is met (monitoring the O-F1 metric and using patience of $50$ epochs for VOC and $10$ epochs for the remaining datasets). All models are implemented with PyTorch\footnote{\url{http://pytorch.org/}} \cite{paszke2017automatic}. For autoregressive models, we train on two variants of annotations: (1) we keep the dataset order (e. g. $\mathrm{LSTM}$ and $\mathrm{TF}$), and (2) we randomly shuffle the labels each time we load an image (e. g. $\mathrm{LSTM_{shuffle}}$ and $\mathrm{TF_{shuffle}}$). 

\subsection{Hyperband details}
The operation of \textsc{Hyperband} is controlled by two hyperparaments. One, $\eta$, controls the aggressiveness of the algorithm, by specifying the ratio of configurations that are kept after each round (the best $1/\eta$ hyperparameters are kept). The other, $R$, controls the maximum number of resources (e.g., epochs) that can be spent by configuration, as well as the total number of configurations to evaluate.  In our experiments, we used $\eta=3$ and $R=600$, where each resource unit is equivalent to $0.15$ training epochs for most datasets ($0.2$ epochs for VOC)\footnote{As VOC is a smaller dataset, we set a larger maximum number of epochs to allow more gradient updates to all models.}, rounding up when necessary. This translates to $~410$ hyperparameter configurations evaluated per model, and a maximum budget of $3\,200$ epochs ($4\,400$ for VOC) for the complete tuning process (with at most $90$ training epochs per model).

For hyperparameter tuning, we allowed \textsc{Hyperband} to sample values from a set of mutually independent categorical distributions, one for each hyperparameter. The hyperparameters considered for all models, and their possible values, are shown in Tables  \ref{tab:hyperparams_common} and \ref{tab:hyperparams_unique}. The hyperparameter values corresponding to the best models found by \textsc{Hyperband} for each dataset are shown in Tables \ref{tab:hyperparams_voc}-\ref{tab:hyperparams_recipe1m}.

\begin{table*}[]
\centering
\begin{tabular}{@{}lc@{}}
\toprule
  \textbf{Hyperparameter} & \textbf{Values} \\ \midrule
  Embedding size & \makecell{$[256, 512, 1\,024, 2\,048]$} \\ 
  Learning rate & \makecell{$[10^{-4}, 10^{-3}, 10^{-2}]$} \\
  Image encoder's learning rate scale & \makecell{$[10^{-2}, 10^{-1}]$} \\ 
  Dropout rate & \makecell{$[0, 0.1, 0.3, 0.5]$} \\ 
  Weight decay & \makecell{$[0, 10^{-4}]$} \\ 
  \bottomrule
\end{tabular}
\caption{Hyperparameters common to all models and their possible values.}
\label{tab:hyperparams_common}
\vspace{-0.5cm}
\end{table*}

\begin{table*}[]
\centering
\resizebox{\textwidth}{!}{
\begin{tabular}{@{}l|cccccc@{}}
\toprule
  \textbf{Models} & $L_t$ & $L_f$ & $n_{att}$  & $\lambda_C$ & $\lambda_{eos}$  \\ \midrule
  $\mathrm{TF}$, $\mathrm{TF}_{\mathrm{shuffle}}$ & \makecell{$[1, 2, 3]$} & \makecell{$-$} & \makecell{$[2, 4, 8]$} & \makecell{$-$} & \makecell{$-$}  \\ \midrule
  
  $\mathrm{TF}_{\mathrm{set}}$  & \makecell{$[1, 2, 3]$} & \makecell{$-$} & \makecell{$[2, 4, 8]$} & \makecell{$-$} & \makecell{$[10^{-3}, 10^{-2}, 10^{-1}, 0.5, 1, 10, 100]$}  \\ \midrule
  
  $\mathrm{LSTM}_{\mathrm{set}}$ & \makecell{$-$} & \makecell{$-$} & \makecell{$-$} & \makecell{$-$} & \makecell{$[10^{-3}, 10^{-2}, 10^{-1}, 0.5, 1, 10, 100]$}  \\ \midrule
  
  $\mathrm{FF}_{\mathrm{BCE}}$, $\mathrm{FF}_{\mathrm{TD}}$, $\mathrm{FF}_{\mathrm{sIoU}}$ & \makecell{$-$} & \makecell{$[0, 1, 2, 3]$} & \makecell{$-$} & \makecell{$-$} & \makecell{$-$}  \\ \midrule
  
  $\mathrm{FF}_{\mathrm{BCE,DC}}$ & \makecell{$-$} & \makecell{$[0, 1, 2, 3]$} & \makecell{$-$} & \makecell{$1$} & \makecell{$-$}  \\ \midrule
  
  $\mathrm{FF}_{\mathrm{BCE,C}}$, $\mathrm{FF}_{\mathrm{TD,C}}$, $\mathrm{FF}_{\mathrm{sIoU,C}}$  & \makecell{$-$} & \makecell{$[0, 1, 2, 3]$} & \makecell{$-$} &\makecell{$[10^{-3}, 10^{-2}, 10^{-1}, 0.5, 1, 10, 100]$} & \makecell{$-$} \\ 
  
 \bottomrule
\end{tabular}}
\caption{Model-specific hyperparameters and their possible values. Models not shown do not have any additional hyperparameters besides those in Table~\ref{tab:hyperparams_common}. $L_t$ and $L_f$ represent the number of transformer layers and fully connected layers, respectively, while $n_{att}$ represents the number of attention heads, $\lambda_C$ represents the weight for the cardinality loss, and $\lambda_{eos}$ represents the weight applied to the end of sequence loss. }
\label{tab:hyperparams_unique}
\vspace{-0.5cm}
\end{table*}

\begin{table*}[]
\centering
\resizebox{\textwidth}{!}{
\begin{tabular}{@{}l|l|ccccccccccccc@{}}
\toprule
  \textbf{Model} & \textbf{Backbone} & $L_t$ & $L_f$ & embedding size & $n_{att}$  & $l_r$ & $\lambda_C$ & $\lambda_{eos}$ & scale & dropout rate & weight decay & total \# of parameters \\ \midrule

$\mathrm{TF}$ & ResNet-50 & \makecell{$3$} & \makecell{$-$} & \makecell{$512$} & \makecell{$8$} & \makecell{$10^{-4}$} & \makecell{$-$} & \makecell{$-$} & \makecell{$0.1$} & \makecell{$0.0$} & \makecell{$10^{-4}$} & $32\,469\,589$ \\ \midrule

$\mathrm{TF_{shuffle}}$ & ResNet-50 & \makecell{$1$} & \makecell{$-$} & \makecell{$512$} & \makecell{$8$} & \makecell{$10^{-4}$} & \makecell{$-$} & \makecell{$-$} & \makecell{$0.1$} & \makecell{$0.1$} & \makecell{$10^{-4}$} & $27\,210\,325$ \\ \midrule

$\mathrm{TF_{set}}$ & ResNet-50 & \makecell{$1$} & \makecell{$-$} & \makecell{$512$} & \makecell{$8$} & \makecell{$10^{-4}$} & \makecell{$-$} & \makecell{$0.5$} & \makecell{$0.1$} & \makecell{$0.1$} & \makecell{$10^{-4}$} & $27\,210\,325$ \\ \midrule

$\mathrm{LSTM}$ & ResNet-50 & \makecell{$-$} & \makecell{$-$} & \makecell{$2048$} & \makecell{$-$} & \makecell{$10^{-4}$} & \makecell{$-$} & \makecell{$-$} & \makecell{$0.1$} & \makecell{$0.5$} & \makecell{$10^{-4}$} & $94\,927\,958$ \\ \midrule

$\mathrm{LSTM_{shuffle}}$ & ResNet-50 & \makecell{$-$} & \makecell{$-$} & \makecell{$2048$} & \makecell{$-$} & \makecell{$10^{-3}$} & \makecell{$-$} & \makecell{$-$} & \makecell{$10^{-2}$} & \makecell{$0.5$} & \makecell{$0.0$} & $94\,927\,958$ \\ \midrule

$\mathrm{LSTM_{set}}$ & ResNet-50 & \makecell{$-$} & \makecell{$-$} & \makecell{$1024$} & \makecell{$-$} & \makecell{$10^{-3}$} & \makecell{$-$} & \makecell{$0.1$} & \makecell{$10^{-2}$} & \makecell{$0.0$} & \makecell{$10^{-4}$} & $43\,491\,414$\\ \midrule

$\mathrm{FF_{BCE}}$ & ResNet-50 & \makecell{$-$} & \makecell{$0$} & \makecell{$2048$} & \makecell{$-$} & \makecell{$10^{-3}$} & \makecell{$-$} & \makecell{$-$} & \makecell{$10^{-2}$} & \makecell{$0.5$} & \makecell{$10^{-4}$} & $23\,549\,012$\\ \midrule

$\mathrm{FF_{BCE,C}}$ & ResNet-50 & \makecell{$-$} & \makecell{$1$} & \makecell{$2048$} & \makecell{$-$} & \makecell{$10^{-4}$} & \makecell{$10^{-2}$} & \makecell{$-$} & \makecell{$0.1$} & \makecell{$0.3$} & \makecell{$0.0$} & $27\,761\,755$\\ \midrule

$\mathrm{FF_{BCE,DC}}$ & ResNet-50 & \makecell{$-$} & \makecell{$0$} & \makecell{$2048$} & \makecell{$-$} & \makecell{$10^{-3}$} & \makecell{$-$} & \makecell{$-$} & \makecell{$10^{-2}$} & \makecell{$0.3$} & \makecell{$0.0$} & $23\,563\,355$\\ \midrule

$\mathrm{FF_{sIoU}}$ & ResNet-50 & \makecell{$-$} & \makecell{$2$} & \makecell{$2048$} & \makecell{$-$} & \makecell{$10^{-4}$} & \makecell{$-$} & \makecell{$-$} & \makecell{$0.1$} & \makecell{$0.0$} & \makecell{$10^{-4}$} & $31\,945\,812$\\ \midrule

$\mathrm{FF_{sIoU,C}}$ & ResNet-50 & \makecell{$-$} & \makecell{$2$} & \makecell{$2048$} & \makecell{$-$} & \makecell{$10^{-4}$} & \makecell{$10^{-2}$} & \makecell{$-$} & \makecell{$0.1$} & \makecell{$0.1$} & \makecell{$10^{-4}$} & $31\,960\,155$\\ \midrule

$\mathrm{FF_{TD}}$ & ResNet-50 & \makecell{$-$} & \makecell{$3$} & \makecell{$512$} & \makecell{$-$} & \makecell{$10^{-3}$} & \makecell{$-$} & \makecell{$-$} & \makecell{$10^{-2}$} & \makecell{$0.0$} & \makecell{$0.0$} & $25\,357\,396$\\ \midrule

$\mathrm{FF_{TD,C}}$ & ResNet-50 & \makecell{$-$} & \makecell{$3$} & \makecell{$512$} & \makecell{$-$} & \makecell{$10^{-4}$} & \makecell{$0.1$} & \makecell{$-$} & \makecell{$0.1$} & \makecell{$0.0$} & \makecell{$10^{-4}$}  & $25\,360\,987$ \\ \midrule \midrule

$\mathrm{FF_{BCE}}$ & ResNet-101 & \makecell{$-$} & \makecell{$0$} & \makecell{$2048$} & \makecell{$-$} & \makecell{$10^{-3}$} & \makecell{$-$} & \makecell{$-$} & \makecell{$10^{-2}$} & \makecell{$0.1$} & \makecell{$10^{-4}$} & $42\,541\,140$ \\ \midrule

$\mathrm{LSTM}$ & ResNet-101 & \makecell{$-$} & \makecell{$-$} & \makecell{$2048$} & \makecell{$-$} & \makecell{$10^{-3}$} & \makecell{$-$} & \makecell{$-$} & \makecell{$10^{-2}$} & \makecell{$0.3$} & \makecell{$10^{-4}$} & $113\,920\,086$ \\ \midrule \midrule

$\mathrm{FF_{BCE}}$ & ResNeXt-101-32x8d & \makecell{$-$} & \makecell{$2$} & \makecell{$512$} & \makecell{$-$} & \makecell{$10^{-4}$} & \makecell{$-$} & \makecell{$-$} & \makecell{$0.1$} & \makecell{$0.3$} & \makecell{$0.0$} & $88\,328\,532$ \\ \midrule

$\mathrm{LSTM}$ & ResNeXt-101-32x8d & \makecell{$-$} & \makecell{$-$} & \makecell{$1024$} & \makecell{$-$} & \makecell{$10^{-4}$} & \makecell{$-$} & \makecell{$-$} & \makecell{$10^{-2}$} & \makecell{$0.5$} & \makecell{$0.0$} & $106\,725\,718$ \\

 \bottomrule
\end{tabular}}
\caption{\textbf{Hyperparameter values chosen by \textsc{Hyperband} for VOC.} $L_t$ and $L_f$ represent the number of transformer layers and fully connected layers, respectively, while $n_{att}$ represents the number of attention heads, $\lambda_C$ refers to the weight for the cardinality loss, $l_r$ to the learning rate, $\lambda_{eos}$ to the weight applied to the end of sequence loss and scale refers to the ratio between the image encoder's and set predictor's learning rates.}
\label{tab:hyperparams_voc}
\vspace{-0.5cm}
\end{table*}

\begin{table*}[]
\centering
\resizebox{\textwidth}{!}{
\begin{tabular}{@{}l|l|cccccccccccc@{}}
\toprule

\textbf{Model} & Backbone & $L_t$ & $L_f$ & embedding size & $n_{att}$  & $l_r$ & $\lambda_C$ & $\lambda_{eos}$ & scale & dropout rate & weight decay & total \# of parameters \\ \midrule

$\mathrm{TF}$ & ResNet-50 & \makecell{$2$} & \makecell{$-$} & \makecell{$256$} & \makecell{$2$} & \makecell{$10^{-3}$} & \makecell{$-$} & \makecell{$-$} & \makecell{$10^{-2}$} & \makecell{$0.1$} & \makecell{$10^{-4}$} & $25\,394\,065$ \\ \midrule

$\mathrm{TF_{shuffle}}$ & ResNet-50 & \makecell{$2$} & \makecell{$-$} & \makecell{$512$} & \makecell{$4$} & \makecell{$10^{-4}$} & \makecell{$-$} & \makecell{$-$} & \makecell{$0.1$} & \makecell{$0.1$} & \makecell{$0.0$} & $29\,901\,457$ \\ \midrule

$\mathrm{TF_{set}}$ & ResNet-50 & \makecell{$3$} & \makecell{$-$} & \makecell{$512$} & \makecell{$2$} & \makecell{$10^{-4}$} & \makecell{$-$} & \makecell{$0.5$} & \makecell{$0.1$} & \makecell{$0.1$} & \makecell{$0.0$} & $32\,531\,089$ \\ \midrule

$\mathrm{LSTM}$ & ResNet-50 & \makecell{$-$} & \makecell{$-$} & \makecell{$1024$} & \makecell{$-$} & \makecell{$10^{-3}$} & \makecell{$-$} & \makecell{$-$} & \makecell{$10^{-2}$} & \makecell{$0.1$} & \makecell{$10^{-4}$} & $43\,614\,354$ \\ \midrule

$\mathrm{LSTM_{shuffle}}$ & ResNet-50 & \makecell{$-$} & \makecell{$-$} & \makecell{$2048$} & \makecell{$-$} & \makecell{$10^{-4}$} & \makecell{$-$} & \makecell{$-$} & \makecell{$0.1$} & \makecell{$0.1$} & \makecell{$10^{-4}$} & $95\,173\,778$ \\ \midrule

$\mathrm{LSTM_{set}}$ & ResNet-50& \makecell{$-$} & \makecell{$-$} & \makecell{$2048$} & \makecell{$-$} & \makecell{$10^{-4}$} & \makecell{$-$} & \makecell{$0.1$} & \makecell{$0.1$} & \makecell{$0.0$} & \makecell{$0.0$} & $95\,173\,778$ \\ \midrule

$\mathrm{FF_{BCE}}$ & ResNet-50 & \makecell{$-$} & \makecell{$2$} & \makecell{$512$} & \makecell{$-$} & \makecell{$10^{-3}$} & \makecell{$-$} & \makecell{$-$} & \makecell{$10^{-2}$} & \makecell{$0.1$} & \makecell{$0.0$} & $25\,125\,008$ \\ \midrule

$\mathrm{FF_{BCE,C}}$ & ResNet-50  & \makecell{$-$} & \makecell{$3$} & \makecell{$2048$} & \makecell{$-$} & \makecell{$10^{-4}$} & \makecell{$10^{-2}$} & \makecell{$-$} & \makecell{$0.1$} & \makecell{$0.1$} & \makecell{$0.0$}  & $36\,306\,083$ \\ \midrule

$\mathrm{FF_{BCE,DC}}$ & ResNet-50 & \makecell{$-$} & \makecell{$2$} & \makecell{$2048$} & \makecell{$-$} & \makecell{$10^{-3}$} & \makecell{$-$} & \makecell{$-$} & \makecell{$10^{-2}$} & \makecell{$0.5$} & \makecell{$0.0$} & $32\,107\,683$ \\ \midrule

$\mathrm{FF_{sIoU}}$ & ResNet-50 & \makecell{$-$} & \makecell{$1$} & \makecell{$2048$} & \makecell{$-$} & \makecell{$10^{-4}$} & \makecell{$-$} & \makecell{$-$} & \makecell{$10^{-2}$} & \makecell{$0.0$} & \makecell{$0.0$} & $27\,870\,352$ \\ \midrule

$\mathrm{FF_{sIoU,C}}$ & ResNet-50 & \makecell{$-$} & \makecell{$2$} & \makecell{$2048$} & \makecell{$-$} & \makecell{$10^{-4}$} & \makecell{$10^{-2}$} & \makecell{$-$} & \makecell{$10^{-2}$} & \makecell{$0.0$} & \makecell{$10^{-4}$} & $32\,107\,683$ \\ \midrule

$\mathrm{FF_{TD,C}}$ & ResNet-50 & \makecell{$-$} & \makecell{$2$} & \makecell{$1024$} & \makecell{$-$} & \makecell{$10^{-3}$} & \makecell{$0.1$} & \makecell{$-$} & \makecell{$10^{-2}$} & \makecell{$0.1$} & \makecell{$10^{-4}$}  & $27\,809\,955$ \\ \midrule \midrule

$\mathrm{FF_{BCE}}$ & ResNet-101 & \makecell{$-$} & \makecell{$1$} & \makecell{$2048$} & \makecell{$-$} & \makecell{$10^{-4}$} & \makecell{$-$} & \makecell{$-$} & \makecell{$0.1$} & \makecell{$0.0$} & \makecell{$0.0$} & $46\,862\,480$ \\ \midrule

$\mathrm{LSTM}$ & ResNet-101 & \makecell{$-$} & \makecell{$-$} & \makecell{$512$} & \makecell{$-$} & \makecell{$10^{-3}$} & \makecell{$-$} & \makecell{$-$} & \makecell{$10^{-2}$} & \makecell{$0.0$} & \makecell{$10^{-4}$} & $48\,096\,914$ \\ \midrule \midrule

$\mathrm{FF_{BCE}}$ & ResNeXt-101-32x8d & \makecell{$-$} & \makecell{$2$} & \makecell{$2048$} & \makecell{$-$} & \makecell{$10^{-4}$} & \makecell{$-$} & \makecell{$-$} & \makecell{$0.1$} & \makecell{$0.0$} & \makecell{$0.0$} & $95\,303\,056$ \\ \midrule

$\mathrm{LSTM}$ & ResNeXt-101-32x8d & \makecell{$-$} & \makecell{$-$} & \makecell{$512$} & \makecell{$-$} & \makecell{$10^{-3}$} & \makecell{$-$} & \makecell{$-$} & \makecell{$10^{-2}$} & \makecell{$0.1$} & \makecell{$10^{-4}$} & $92\,339\,090$ \\

 \bottomrule
\end{tabular}}
\caption{\textbf{Hyperparameter values chosen by \textsc{Hyperband} for COCO.} $L_t$ and $L_f$ represent the number of transformer layers and fully connected layers, respectively, while $n_{att}$ represents the number of attention heads, $\lambda_C$ refers to the weight for the cardinality loss, $l_r$ to the learning rate, $\lambda_{eos}$ to the weight applied to the end of sequence loss and scale refers to the ratio between the image encoder's and set predictor's learning rates.}
\label{tab:hyperparams_coco}
\vspace{-0.5cm}
\end{table*}

\begin{table*}[]
\centering
\resizebox{\textwidth}{!}{
\begin{tabular}{@{}l|l|ccccccccccccc@{}}
\toprule

\textbf{Model} & Backbone & $L_t$ & $L_f$ & embedding size & $n_{att}$  & $l_r$ & $\lambda_C$ & $\lambda_{eos}$ & scale & dropout rate & weight decay & total \# of parameters \\ \midrule
  $\mathrm{TF}$ & ResNet-50 & \makecell{$3$} & \makecell{$-$} & \makecell{$256$} & \makecell{$4$} & \makecell{$10^{-4}$} & \makecell{$-$} & \makecell{$-$} & \makecell{$0.1$} & \makecell{$0.0$} & \makecell{$10^{-4}$} & $26\,054\,034$ \\ \midrule

$\mathrm{TF_{shuffle}}$ & ResNet-50 & \makecell{$1$} & \makecell{$-$} & \makecell{$512$} & \makecell{$8$} & \makecell{$10^{-4}$} & \makecell{$-$} & \makecell{$-$} & \makecell{$10^{-2}$} & \makecell{$0.3$} & \makecell{$0.0$} & $27\,272\,850$ \\ \midrule

$\mathrm{TF_{set}}$ & ResNet-50 & \makecell{$1$} & \makecell{$-$} & \makecell{$256$} & \makecell{$4$} & \makecell{$10^{-4}$} & \makecell{$-$} & \makecell{$0.1$} & \makecell{$0.1$} & \makecell{$0.1$} & \makecell{$0.0$} & $24\,735\,122$ \\ \midrule

$\mathrm{LSTM}$ & ResNet-50 & \makecell{$-$} & \makecell{$-$} & \makecell{$2048$} & \makecell{$-$} & \makecell{$10^{-4}$} & \makecell{$-$} & \makecell{$-$} & \makecell{$10^{-2}$} & \makecell{$0.3$} & \makecell{$10^{-4}$} & $95\,177\,875$ \\ \midrule

$\mathrm{LSTM_{shuffle}}$ & ResNet-50 & \makecell{$-$} & \makecell{$-$} & \makecell{$256$} & \makecell{$-$} & \makecell{$10^{-4}$} & \makecell{$-$} & \makecell{$-$} & \makecell{$0.1$} & \makecell{$0.5$} & \makecell{$0.0$} & $25\,192\,851$ \\ \midrule

$\mathrm{LSTM_{set}}$ & ResNet-50 & \makecell{$-$} & \makecell{$-$} & \makecell{$1024$} & \makecell{$-$} & \makecell{$10^{-4}$} & \makecell{$-$} & \makecell{$0.1$} & \makecell{$10^{-2}$} & \makecell{$0.3$} & \makecell{$0.0$} & $43\,616\,403$ \\ \midrule

$\mathrm{FF_{BCE}}$ & ResNet-50 & \makecell{$-$} & \makecell{$1$} & \makecell{$2048$} & \makecell{$-$} & \makecell{$10^{-4}$} & \makecell{$-$} & \makecell{$-$} & \makecell{$0.1$} & \makecell{$0.1$} & \makecell{$0.0$} & $27\,872\,401$ \\ \midrule

$\mathrm{FF_{BCE,C}}$ & ResNet-50 & \makecell{$-$} & \makecell{$1$} & \makecell{$1024$} & \makecell{$-$} & \makecell{$10^{-4}$} & \makecell{$10^{-2}$} & \makecell{$-$} & \makecell{$0.1$} & \makecell{$0.1$} & \makecell{$0.0$} & $26\,754\,206$ \\ \midrule

$\mathrm{FF_{BCE,DC}}$ & ResNet-50 & \makecell{$-$} & \makecell{$1$} & \makecell{$1024$} & \makecell{$-$} & \makecell{$10^{-3}$} & \makecell{$-$} & \makecell{$-$} & \makecell{$10^{-2}$} & \makecell{$0.3$} & \makecell{$0.0$} & $26\,754\,206$ \\ \midrule

$\mathrm{FF_{sIoU}}$ & ResNet-50 & \makecell{$-$} & \makecell{$1$} & \makecell{$2048$} & \makecell{$-$} & \makecell{$10^{-4}$} & \makecell{$-$} & \makecell{$-$} & \makecell{$0.1$} & \makecell{$0.1$} & \makecell{$0.0$} & $27\,872\,401$ \\ \midrule

$\mathrm{FF_{sIoU,C}}$ & ResNet-50 & \makecell{$-$} & \makecell{$1$} & \makecell{$1024$} & \makecell{$-$} & \makecell{$10^{-4}$} & \makecell{$10^{-2}$} & \makecell{$-$} & \makecell{$0.1$} & \makecell{$0.0$} & \makecell{$0.0$}  & $26\,754\,206$ \\ \midrule

$\mathrm{FF_{TD,C}}$ & ResNet-50 & \makecell{$-$} & \makecell{$2$} & \makecell{$2048$} & \makecell{$-$} & \makecell{$10^{-4}$} & \makecell{$0.5$} & \makecell{$-$} & \makecell{$0.1$} & \makecell{$0.0$} & \makecell{$10^{-4}$} & $32\,097\,438$ \\ \midrule \midrule

$\mathrm{FF_{BCE}}$ & ResNet-101 & \makecell{$-$} & \makecell{$1$} & \makecell{$2048$} & \makecell{$-$} & \makecell{$10^{-4}$} & \makecell{$-$} & \makecell{$-$} & \makecell{$10^{-2}$} & \makecell{$0.3$} & \makecell{$0.0$} & $46\,864\,529$ \\ \midrule

$\mathrm{LSTM}$ & ResNet-101 & \makecell{$-$} & \makecell{$-$} & \makecell{$2048$} & \makecell{$-$} & \makecell{$10^{-4}$} & \makecell{$-$} & \makecell{$-$} & \makecell{$10^{-2}$} & \makecell{$0.0$} & \makecell{$10^{-4}$} & $114\,170\,003$ \\ \midrule \midrule

$\mathrm{FF_{BCE}}$ & ResNeXt-101-32x8d & \makecell{$-$} & \makecell{$1$} & \makecell{$2048$} & \makecell{$-$} & \makecell{$10^{-4}$} & \makecell{$-$} & \makecell{$-$} & \makecell{$10^{-2}$} & \makecell{$0.3$} & \makecell{$0.0$} & $91\,106\,705$ \\ \midrule

$\mathrm{LSTM}$ & ResNeXt-101-32x8d & \makecell{$-$} & \makecell{$-$} & \makecell{$1024$} & \makecell{$-$} & \makecell{$10^{-4}$} & \makecell{$-$} & \makecell{$-$} & \makecell{$10^{-2}$} & \makecell{$0.0$} & \makecell{$0.0$} & $106\,850\,707$ \\

 \bottomrule
\end{tabular}}
\caption{\textbf{Hyperparameter values chosen by \textsc{Hyperband} for NUS-WIDE.} $L_t$ and $L_f$ represent the number of transformer layers and fully connected layers, respectively, while $n_{att}$ represents the number of attention heads, $\lambda_C$ refers to the weight for the cardinality loss, $l_r$ to the learning rate, $\lambda_{eos}$ to the weight applied to the end of sequence loss and scale refers to the ratio between the image encoder's and set predictor's learning rates.}
\label{tab:hyperparams_nuswide}
\vspace{-0.5cm}
\end{table*}

\begin{table*}[]
\centering
\resizebox{\textwidth}{!}{
\begin{tabular}{@{}l|l|ccccccccccccc@{}}
\toprule
  \textbf{Model} & Backbone & $L_t$ & $L_f$ & embedding size & $n_{att}$  & $l_r$ & $\lambda_C$ & $\lambda_{eos}$ & scale & dropout rate & weight decay & total \# of parameters \\ \midrule
  
$\mathrm{TF}$ & ResNet-50 & \makecell{$1$} & \makecell{$-$} & \makecell{$256$} & \makecell{$4$} & \makecell{$10^{-3}$} & \makecell{$-$} & \makecell{$-$} & \makecell{$0.1$} & \makecell{$0.1$} & \makecell{$10^{-4}$} & $24\,770\,519$ \\ \midrule

$\mathrm{TF_{shuffle}}$ & ResNet-50 & \makecell{$2$} & \makecell{$-$} & \makecell{$256$} & \makecell{$8$} & \makecell{$10^{-3}$} & \makecell{$-$} & \makecell{$-$} & \makecell{$0.1$} & \makecell{$0.1$} & \makecell{$0.0$} & $25\,429\,975$ \\ \midrule

$\mathrm{TF_{set}}$ & ResNet-50 & \makecell{$2$} & \makecell{$-$} & \makecell{$1024$} & \makecell{$8$} & \makecell{$10^{-4}$} & \makecell{$-$} & \makecell{$0.1$} & \makecell{$0.1$} & \makecell{$0.1$} & \makecell{$10^{-4}$} & $46\,923\,991$ \\ \midrule

$\mathrm{LSTM}$ & ResNet-50 & \makecell{$-$} & \makecell{$-$} & \makecell{$2048$} & \makecell{$-$} & \makecell{$10^{-3}$} & \makecell{$-$} & \makecell{$-$} & \makecell{$10^{-2}$} & \makecell{$0.3$} & \makecell{$10^{-4}$} & $95\,460\,568$ \\ \midrule

$\mathrm{LSTM_{shuffle}}$ & ResNet-50 & \makecell{$-$} & \makecell{$-$} & \makecell{$512$} & \makecell{$-$} & \makecell{$10^{-3}$} & \makecell{$-$} & \makecell{$-$} & \makecell{$0.1$} & \makecell{$0.0$} & \makecell{$0.0$} & $29\,176\,536$ \\ \midrule

$\mathrm{LSTM_{set}}$ & ResNet-50 & \makecell{$-$} & \makecell{$-$} & \makecell{$1024$} & \makecell{$-$} & \makecell{$10^{-3}$} & \makecell{$-$} & \makecell{$0.1$} & \makecell{$0.1$} & \makecell{$0.0$} & \makecell{$0.0$} & $43\,757\,784$ \\ \midrule

$\mathrm{FF_{BCE}}$ & ResNet-50 & \makecell{$-$} & \makecell{$0$} & \makecell{$1024$} & \makecell{$-$} & \makecell{$10^{-2}$} & \makecell{$-$} & \makecell{$-$} & \makecell{$10^{-2}$} & \makecell{$0.1$} & \makecell{$0.0$} & $25\,760\,982$ \\ \midrule

$\mathrm{FF_{BCE,C}}$ & ResNet-50 & \makecell{$-$} & \makecell{$0$} & \makecell{$2048$} & \makecell{$-$} & \makecell{$10^{-3}$} & \makecell{$10^{-2}$} & \makecell{$-$} & \makecell{$10^{-2}$} & \makecell{$0.1$} & \makecell{$10^{-4}$} & $23\,878\,901$ \\ \midrule

$\mathrm{FF_{BCE,DC}}$ & ResNet-50 & \makecell{$-$} & \makecell{$3$} & \makecell{$2048$} & \makecell{$-$} & \makecell{$10^{-4}$} & \makecell{$-$} & \makecell{$-$} & \makecell{$0.1$} & \makecell{$0.1$} & \makecell{$0.0$} & $36\,474\,101$ \\ \midrule

$\mathrm{FF_{sIoU}}$ & ResNet-50 & \makecell{$-$} & \makecell{$1$} & \makecell{$2048$} & \makecell{$-$} & \makecell{$10^{-4}$} & \makecell{$-$} & \makecell{$-$} & \makecell{$0.1$} & \makecell{$0.0$} & \makecell{$0.0$} & $28\,013\,782$ \\ \midrule

$\mathrm{FF_{sIoU,C}}$ & ResNet-50 & \makecell{$-$} & \makecell{$1$} & \makecell{$2048$} & \makecell{$-$} & \makecell{$10^{-4}$} & \makecell{$10^{-3}$} & \makecell{$-$} & \makecell{$0.1$} & \makecell{$0.0$} & \makecell{$0.0$} & $28\,077\,301$ \\ \midrule

$\mathrm{FF_{TD}}$ & ResNet-50 & \makecell{$-$} & \makecell{$1$} & \makecell{$2048$} & \makecell{$-$} & \makecell{$10^{-2}$} & \makecell{$-$} & \makecell{$-$} & \makecell{$10^{-2}$} & \makecell{$0.1$} & \makecell{$0.0$} & $28\,013\,782$ \\ \midrule

$\mathrm{FF_{TD,C}}$ & ResNet-50 & \makecell{$-$} & \makecell{$2$} & \makecell{$2048$} & \makecell{$-$} & \makecell{$10^{-3}$} & \makecell{$10^{-2}$} & \makecell{$-$} & \makecell{$0.1$} & \makecell{$0.1$} & \makecell{$0.0$} & $32\,275\,701$ \\ \midrule \midrule

$\mathrm{FF_{BCE}}$ & ResNet-101 & \makecell{$-$} & \makecell{$1$} & \makecell{$2048$} & \makecell{$-$} & \makecell{$10^{-4}$} & \makecell{$-$} & \makecell{$-$} & \makecell{$0.1$} & \makecell{$0.0$} & \makecell{$0.0$} & $47\,005\,910$ \\ \midrule

$\mathrm{LSTM}$ & ResNet-101 & \makecell{$-$} & \makecell{$-$} & \makecell{$512$} & \makecell{$-$} & \makecell{$10^{-3}$} & \makecell{$-$} & \makecell{$-$} & \makecell{$10^{-2}$} & \makecell{$0.0$} & \makecell{$10^{-4}$} & $48\,168\,664$ \\ \midrule \midrule

$\mathrm{FF_{BCE}}$ & ResNeXt-101-32x8d & \makecell{$-$} & \makecell{$0$} & \makecell{$2048$} & \makecell{$-$} & \makecell{$10^{-3}$} & \makecell{$-$} & \makecell{$-$} & \makecell{$10^{-2}$} & \makecell{$0.0$} & \makecell{$10^{-4}$} & $87\,049\,686$ \\ \midrule

$\mathrm{LSTM}$ & ResNeXt-101-32x8d & \makecell{$-$} & \makecell{$-$} & \makecell{$256$} & \makecell{$-$} & \makecell{$10^{-3}$} & \makecell{$-$} & \makecell{$-$} & \makecell{$10^{-2}$} & \makecell{$0.1$} & \makecell{$10^{-4}$} & $88\,462\,552$ \\

 \bottomrule
\end{tabular}}
\caption{\textbf{Hyperparameter values chosen by \textsc{Hyperband} for ADE20k.} $L_t$ and $L_f$ represent the number of transformer layers and fully connected layers, respectively, while $n_{att}$ represents the number of attention heads, $\lambda_C$ refers to the weight for the cardinality loss, $l_r$ to the learning rate, $\lambda_{eos}$ to the weight applied to the end of sequence loss and scale refers to the ratio between the image encoder's and set predictor's learning rates.}
\label{tab:hyperparams_ade20k}
\vspace{-0.5cm}
\end{table*}

\begin{table*}[]
\centering
\resizebox{\textwidth}{!}{
\begin{tabular}{@{}l|l|ccccccccccccc@{}}
\toprule
  \textbf{Model} & Backbone & $L_t$ & $L_f$ & embedding size & $n_{att}$  & $l_r$ & $\lambda_C$ & $\lambda_{eos}$ & scale & dropout rate & weight decay & total \# of parameters \\ \midrule
 
$\mathrm{TF}$ & ResNet-50 & \makecell{$1$} & \makecell{$-$} & \makecell{$2048$} & \makecell{$8$} & \makecell{$10^{-4}$} & \makecell{$-$} & \makecell{$-$} & \makecell{$0.1$} & \makecell{$0.3$} & \makecell{$10^{-4}$} & $71\,582\,223$ \\ \midrule

$\mathrm{TF_{shuffle}}$ & ResNet-50 & \makecell{$2$} & \makecell{$-$} & \makecell{$256$} & \makecell{$4$} & \makecell{$10^{-3}$} & \makecell{$-$} & \makecell{$-$} & \makecell{$0.1$} & \makecell{$0.1$} & \makecell{$0.0$} & $26\,115\,343$ \\ \midrule

$\mathrm{TF_{set}}$ & ResNet-50 & \makecell{$2$} & \makecell{$-$} & \makecell{$256$} & \makecell{$2$} & \makecell{$10^{-3}$} & \makecell{$-$} & \makecell{$10^{-3}$} & \makecell{$0.1$} & \makecell{$0.1$} & \makecell{$0.0$} & $26\,115\,343$ \\ \midrule

$\mathrm{LSTM}$ & ResNet-50 & \makecell{$-$} & \makecell{$-$} & \makecell{$2048$} & \makecell{$-$} & \makecell{$10^{-4}$} & \makecell{$-$} & \makecell{$-$} & \makecell{$0.1$} & \makecell{$0.5$} & \makecell{$10^{-4}$} & $100\,934\,160$ \\ \midrule

$\mathrm{LSTM_{shuffle}}$ & ResNet-50 & \makecell{$-$} & \makecell{$-$} & \makecell{$2048$} & \makecell{$-$} & \makecell{$10^{-3}$} & \makecell{$-$} & \makecell{$-$} & \makecell{$0.1$} & \makecell{$0.1$} & \makecell{$10^{-4}$} & $100\,934\,160$ \\ \midrule

$\mathrm{LSTM_{set}}$ & ResNet-50 & \makecell{$-$} & \makecell{$-$} & \makecell{$2048$} & \makecell{$-$} & \makecell{$10^{-4}$} & \makecell{$-$} & \makecell{$10^{-3}$} & \makecell{$0.1$} & \makecell{$0.5$} & \makecell{$0.0$} & $100\,934\,160$ \\ \midrule

$\mathrm{FF_{BCE}}$ & ResNet-50 & \makecell{$-$} & \makecell{$2$} & \makecell{$2048$} & \makecell{$-$} & \makecell{$10^{-3}$} & \makecell{$-$} & \makecell{$-$} & \makecell{$10^{-2}$} & \makecell{$0.0$} & \makecell{$0.0$} & $34\,949\,646$ \\ \midrule

$\mathrm{FF_{BCE,C}}$ & ResNet-50 & \makecell{$-$} & \makecell{$3$} & \makecell{$2048$} & \makecell{$-$} & \makecell{$10^{-3}$} & \makecell{$10^{-3}$} & \makecell{$-$} & \makecell{$10^{-2}$} & \makecell{$0.1$} & \makecell{$0.0$} & $39\,189\,026$ \\ \midrule

$\mathrm{FF_{BCE,DC}}$ & ResNet-50 & \makecell{$-$} & \makecell{$2$} & \makecell{$1024$} & \makecell{$-$} & \makecell{$10^{-3}$} & \makecell{$-$} & \makecell{$-$} & \makecell{$10^{-2}$} & \makecell{$0.3$} & \makecell{$0.0$} & $29\,252\,130$ \\ \midrule

$\mathrm{FF_{sIoU}}$ & ResNet-50 & \makecell{$-$} & \makecell{$1$} & \makecell{$2048$} & \makecell{$-$} & \makecell{$10^{-4}$} & \makecell{$-$} & \makecell{$-$} & \makecell{$0.1$} & \makecell{$0.1$} & \makecell{$0.0$} & $30\,751\,246$ \\ \midrule

$\mathrm{FF_{sIoU,C}}$ & ResNet-50 & \makecell{$-$} & \makecell{$1$} & \makecell{$1024$} & \makecell{$-$} & \makecell{$10^{-3}$} & \makecell{$0.1$} & \makecell{$-$} & \makecell{$10^{-2}$} & \makecell{$0.0$} & \makecell{$0.0$} & $28\,201\,506$ \\ \midrule

$\mathrm{FF_{TD}}$ & ResNet-50 & \makecell{$-$} & \makecell{$3$} & \makecell{$1024$} & \makecell{$-$} & \makecell{$10^{-4}$} & \makecell{$-$} & \makecell{$-$} & \makecell{$0.1$} & \makecell{$0.0$} & \makecell{$10^{-4}$} & $30\,282\,254$ \\ \midrule

$\mathrm{FF_{TD,C}}$ & ResNet-50 & \makecell{$-$} & \makecell{$2$} & \makecell{$2048$} & \makecell{$-$} & \makecell{$10^{-3}$} & \makecell{$10^{-3}$} & \makecell{$-$} & \makecell{$0.1$} & \makecell{$0.0$} & \makecell{$0.0$} & $34\,990\,626$ \\ \midrule \midrule

$\mathrm{FF_{BCE}}$ & ResNet-101 & \makecell{$-$} & \makecell{$0$} & \makecell{$512$} & \makecell{$-$} & \makecell{$10^{-3}$} & \makecell{$-$} & \makecell{$-$} & \makecell{$0.1$} & \makecell{$0.3$} & \makecell{$0.0$} & $44\,312\,078$ \\ \midrule

$\mathrm{LSTM}$ & ResNet-101 & \makecell{$-$} & \makecell{$-$} & \makecell{$2048$} & \makecell{$-$} & \makecell{$10^{-4}$} & \makecell{$-$} & \makecell{$-$} & \makecell{$0.1$} & \makecell{$0.5$} & \makecell{$10^{-4}$} & $119\,926\,288$ \\ \midrule\midrule

$\mathrm{FF_{BCE}}$ & ResNeXt-101-32x8d & \makecell{$-$} & \makecell{$3$} & \makecell{$2048$} & \makecell{$-$} & \makecell{$10^{-4}$} & \makecell{$-$} & \makecell{$-$} & \makecell{$0.1$} & \makecell{$0.0$} & \makecell{$0.0$} & $102\,382\,350$ \\ \midrule

$\mathrm{LSTM}$ & ResNeXt-101-32x8d & \makecell{$-$} & \makecell{$-$} & \makecell{$1024$} & \makecell{$-$} & \makecell{$10^{-4}$} & \makecell{$-$} & \makecell{$-$} & \makecell{$0.1$} & \makecell{$0.1$} & \makecell{$10^{-4}$} & $109\,729\,552$ \\ 

  \bottomrule
\end{tabular}}
\caption{\textbf{Hyperparameter values chosen by \textsc{Hyperband} for Recipe1M.} $L_t$ and $L_f$ represent the number of transformer layers and fully connected layers, respectively, while $n_{att}$ represents the number of attention heads, $\lambda_C$ refers to the weight for the cardinality loss, $l_r$ to the learning rate, $\lambda_{eos}$ to the weight applied to the end of sequence loss and scale refers to the ratio between the image encoder's and set predictor's learning rates.}
\label{tab:hyperparams_recipe1m}
\vspace{-0.5cm}
\end{table*}

\end{document}